\documentclass[11pt, hidelinks]{article} 

\usepackage{hyperref}
\usepackage{url}
\usepackage{mathtools}
\usepackage{fullpage}

\usepackage{amsmath,amssymb,amsthm, amsfonts}
\usepackage{mathbbol}
\usepackage{mathrsfs}  
\usepackage{multicol}
\usepackage{rotating}
\usepackage{enumitem, comment, xifthen}

\usepackage{xcolor}
\usepackage{mathbbol}

\usepackage{booktabs}

\theoremstyle{plain}
\newtheorem{theorem}{Theorem}
\newtheorem{lemma}{Lemma}
\newtheorem{corollary}{Corollary}
\newtheorem{proposition}{Proposition}

\theoremstyle{definition}

\theoremstyle{remark}

\usepackage[framemethod=TikZ]{mdframed}

\mdfsetup{%
  roundcorner = 20pt
  backgroundcolor=black!10
}

\newcommand{\myframe}[1]{
\begin{mdframed}[backgroundcolor=black!10, roundcorner=5pt]
  #1
\end{mdframed}
}

\newcommand{\Vnoise}{\ensuremath{W}}
\newcommand{\VnoiseIt}[1]{\ITCOMMAND{\Vnoise}{#1}}

\newcommand{\xstate}[1]{\ensuremath{x_{#1}}}

\newcommand{\myoperator}[1]{\mathcal{#1}}

\newcommand{\ITER}[2]{#1_{#2}}

\newcommand{\ITCOMMAND}[2]{\ensuremath{#1}_{#2}}
\newcommand{\pathit}[1]{\ITCOMMAND{P}{#1}}
\newcommand{\pathitprime}[1]{\ITCOMMAND{P'}{#1}}

\newcommand{\onevec}{\ensuremath{\mathbb{1}}}

\newcommand{\thetait}[1]{\ITCOMMAND{\theta}{#1}}
\newcommand{\DeltaIt}[1]{\ITCOMMAND{\Delta}{#1}}
\newcommand{\DelIt}[1]{\DeltaIt{#1}}
\newcommand{\wnoiseit}[1]{\ITCOMMAND{W}{#1}}

\newcommand{\Hop}{\ensuremath{\mathcal{H}}}

\newcommand{\titer}{\ensuremath{k}}

\newcommand{\state}{\ensuremath{x}}
\newcommand{\action}{\ensuremath{u}}

\newcommand{\reward}{\ensuremath{r}}

\newcommand{\TfunPlain}{\ensuremath{\mathbb{P}}}
\newcommand{\Trans}[3]{\ensuremath{\TfunPlain_{#2}(#3 \mid #1)}}
\newcommand{\Tfun}{\Trans}

\newcommand{\discount}{\ensuremath{\gamma}}

\newcommand{\Bellman}{\ensuremath{\myoperator{T}}}

\newcommand{\EmpBellman}{\ensuremath{\widehat{\Bellman}}}
\newcommand{\EmpBellmanIter}[1]{\ITCOMMAND{\EmpBellman}{#1}}

\newcommand{\pol}{\ensuremath{\pi}}

\newcommand{\polopt}{\ensuremath{\pol^*}}

\newcommand{\ActionSpace}{\ensuremath{\mathcal{U}}}
\newcommand{\StateSpace}{\ensuremath{\mathcal{X}}}

\newcommand{\iter}{\ensuremath{\ell}}

\newcommand{\stepit}[1]{\ensuremath{\stepcon_{#1}}}

\newcommand{\stepcon}{\ensuremath{\lambda}}

\newcommand{\bcar}{\begin{itemize}}
\newcommand{\ecar}{\end{itemize}}

\newcommand{\thetabar}{\widebar{\theta}}

\newcommand{\thetastar}{\ensuremath{{\theta^*}}}

\newcommand{\Bou}{\ensuremath{B}}
\newcommand{\bou}{\ensuremath{b}}

\newcommand{\thetahat}{\ensuremath{\widehat{\theta}}}

\newcommand{\vsmall}{\vspace*{.1in}}

\definecolor{MyGray}{rgb}{0.9,0.9,0.9}

\makeatletter\newenvironment{graybox}{ 
\begin{lrbox}{\@tempboxa}
\begin{minipage}{0.985\columnwidth}}{\end{minipage}
\end{lrbox}%
\colorbox{MyGray}{\usebox{\@tempboxa}} }
\makeatother



\newcommand{\Zback}[1]{\ensuremath{Z^{\backslash i}}}

\newcommand{\xsamstack}[1]{\ensuremath{x_1^\numobs}}
\newcommand{\Xsamstack}[1]{\ensuremath{X_1^\numobs}}

\newcommand{\widgraph}[2]{\includegraphics[keepaspectratio,width=#1]{#2}}

\newcommand{\fancysoln}[1]{
\ifthenelse{\equal{\doctype}{WITHSOLS}}
{
\begin{soln}
#1
\end{soln}
}
{
}
}

\newcommand{\order}{\ensuremath{\mathcal{O}}}

\long\def\comment#1{}

\makeatletter
\def\@cite#1#2{[\if@tempswa #2 \fi #1]}
\makeatother

\newcommand{\defn}{\vcentcolon=}

\newcommand{\var}{\ensuremath{\operatorname{var}}}

\newcommand{\real}{\ensuremath{\mathbb{R}}}

\newcommand{\numobs}{\ensuremath{n}}

\newcommand{\mprob}{\ensuremath{\mathbb{P}}}

\newcommand{\mymathbf}[1]{\ensuremath{\mathbf{#1}}}

\newcommand{\Pmat}{\ensuremath{\mymathbf{P}}}

\newlength{\widebarargwidth}
\newlength{\widebarargheight}
\newlength{\widebarargdepth}
\DeclareRobustCommand{\widebar}[1]{%
  \settowidth{\widebarargwidth}{\ensuremath{#1}}%
  \settoheight{\widebarargheight}{\ensuremath{#1}}%
  \settodepth{\widebarargdepth}{\ensuremath{#1}}%
  \addtolength{\widebarargwidth}{-0.3\widebarargheight}%
  \addtolength{\widebarargwidth}{-0.3\widebarargdepth}%
  \makebox[0pt][l]{\hspace{0.3\widebarargheight}%
    \hspace{0.3\widebarargdepth}%
    \addtolength{\widebarargheight}{0.3ex}%
    \rule[\widebarargheight]{0.95\widebarargwidth}{0.1ex}}%
  {#1}}

\newcommand{\Exs}{\ensuremath{\mathbb{E}}}

\newcommand{\figdir}{.}

\makeatletter
\long\def\@makecaption#1#2{
        \vskip 0.8ex
        \setbox\@tempboxa\hbox{\small {\bf #1:} #2}
        \parindent 1.5em  
        \dimen0=\hsize
        \advance\dimen0 by -3em
        \ifdim \wd\@tempboxa >\dimen0
                \hbox to \hsize{
                        \parindent 0em
                        \hfil 
                        \parbox{\dimen0}{\def\baselinestretch{0.96}\small
                                {\bf #1.} #2
                                } 
                        \hfil}
        \else \hbox to \hsize{\hfil \box\@tempboxa \hfil}
        \fi
        }
\makeatother

\newcommand{\unicon}{\ensuremath{c}}

\newcommand{\rmax}{\ensuremath{r_{\mbox{\tiny{max}}}}}

\newcommand{\Qpol}{\ensuremath{\theta^\pol}}

\newcommand{\IdMat}{\ensuremath{\mathbf{I}}}

\newcommand{\MDP}{\ensuremath{\mathscr{M}}}

\newcommand{\N}{\ensuremath{N}}
\newcommand{\Nit}[1]{\ITER{\N}{#1}}
\newcommand{\thetabarit}[1]{\ITER{\mythetabar}{#1}}

\newcommand{\BellmanMix}{\ensuremath{\widetilde{\Bellman}}}

\newcommand{\VnoiseA}{\ensuremath{\VnoiseIt{\titer}^\prime}}
\newcommand{\VnoiseAIter}[1]{\ITER{\Vnoise^\prime}{#1}}
\newcommand{\VnoiseB}{\ensuremath{\Vnoise^\circ}}
\newcommand{\VnoiseC}{\ensuremath{\Vnoise^\dagger}}

\newcommand{\PnoiseA}{\ensuremath{\pathit{\titer}^\prime}}

\newcommand{\BellmanMixN}{\ensuremath{\BellmanMix_{\N}}}

\newcommand{\Dim}{\ensuremath{D}}

\newcommand{\epochit}{\ensuremath{m}}
\newcommand{\uniconprime}{\ensuremath{\unicon'}}
\newcommand{\BouZero}{\ensuremath{\Bou_0}}

\usepackage[usestackEOL]{stackengine}
\usepackage{scalerel}
\def\myoverline#1{\ThisStyle{%
  \setbox0=\hbox{$\SavedStyle#1$}%
  \stackengine{0.6\LMpt}{$\SavedStyle#1$}{\rule{\wd0}{0.8\LMpt}}{O}{c}{F}{F}{S}%
}}

\newcommand{\mythetabar}{\myoverline{\theta}}

\newcommand{\EpLength}{\ensuremath{K}}

\newcommand{\NumEpoch}{\ensuremath{M}}

\newcommand{\eiter}{\ensuremath{m}}
\newcommand{\HopTil}{\ensuremath{\widetilde{\Hop}}}
\newcommand{\HopTilN}{\ensuremath{\HopTil_{\N}}}
\newcommand{\HopK}{\ensuremath{\widehat{\Hop}_{\titer}}}

\newcommand{\Nsamp}{\ensuremath{T}}

\newcommand{\VARUP}{\ensuremath{\mathcal{V}}}

\newcommand{\SamN}{\ensuremath{\mathfrak{D}_\N}}

\renewcommand{\Bou}{\ensuremath{b}}

\newcommand{\HopJ}{\ensuremath{\mathcal{J}}}
\newcommand{\HopJEmp}{\ensuremath{\widehat{\HopJ}}}
\newcommand{\HopJEmpIter}[1]{\ITER{\HopJEmp}{#1}}
\newcommand{\Eff}{\ensuremath{E}}

\newcommand{\rtil}{\ensuremath{\widetilde{\reward}}}

\newcommand{\PhatPol}{\ensuremath{\Pmat^{\polhat}}}
\newcommand{\PPol}{\ensuremath{\Pmat^{\pol^*}}}
\newcommand{\polhat}{\ensuremath{\hat{\pol}}}
\newcommand{\LOGDNSQ}{\frac{\log(8 \NumEpoch \Dim/\delta)}{\N}}
\newcommand{\LOGDN}{\sqrt{\frac{\log(8 \NumEpoch \Dim/\delta)}{\N}}}

\newcommand{\boldpara}[1]{\noindent {\bf{#1:}} }
\newcommand{\maxplus}[1]{\ensuremath{|#1|_+}}
\newcommand{\Tmmax}{\ensuremath{T_{\mbox{\tiny{max}}}}}

\begin{document}

\begin{center}
  {\bf{\LARGE{Variance-reduced $Q$-learning is minimax optimal}}}

  \vspace*{0.5in}
  
  \begin{tabular}{c}
    Martin J. Wainwright  \\
    Departments of Statistics and EECS \\
    UC Berkeley \\
    Voleon Group, Berkeley, CA \\
    \texttt{wainwrig@berkeley.edu}
  \end{tabular}
  
  \vsmall
  
  \begin{abstract}
    We introduce and analyze a form of variance-reduced $Q$-learning.
    For $\discount$-discounted MDPs with finite state space
    $\StateSpace$ and action space $\ActionSpace$, we prove that it
    yields an $\epsilon$-accurate estimate of the optimal $Q$-function
    in the $\ell_\infty$-norm using $\mathcal{O} \left(
    \left(\frac{D}{ \epsilon^2 (1-\discount)^3} \right) \; \log \left(
    \frac{D}{(1-\discount)} \right) \right)$ samples, where $D =
    |\StateSpace| \times |\ActionSpace|$.  This guarantee matches
    known minimax lower bounds up to a logarithmic factor in the
    discount complexity.  By contrast, our past work shows that
    ordinary $Q$-learning has worst-case quartic scaling in the
    discount complexity.
  \end{abstract}

\end{center}


\section{Introduction}

Markov decision processes and reinforcement learning algorithms
provide a flexible framework for decision-making in dynamic settings,
and have been studied for decades (e.g.,~\cite{Puterman05, SutBar18,
  Bertsekas_dyn1, BerTsi96, Sze09}).  Given the explosion in the
amount of available data and computing power, recent years have
witnessed dramatic success of reinforcement learning (RL) techniques
in various application domains (e.g.,~\cite{tobin17, levine15,
  silver16, mnih15, SutBar18}).  Providing a firm theoretical
foundation to the trade-offs intrinsic to different classes of
methods, as characterized by their access to the underlying Markov
decision process, is a major open question in RL.

Such performance trade-offs have been studied in some detail for both
MDPs with finite state-action spaces (e.g.,~\cite{KeaSin99, Sze97,
  EveMan03, Aza11, AgaJia17, AzaMunKap13, AzaOsbMun17, EveMan03,
  Sze97,LatHut14, JinZhuBubJor18, Wai19CC}), as well as for linear
state space models with quadratic rewards, known as the linear
quadratic regulator (LQR) problem
(e.g.,~\cite{abbasi2011,abbasi2018,abeille2017, CohKorMan19,
  ManTuRec19, TuRec19, FazGeKakMes18, MalPanBhaKhaBarWai19}).  While
both classes of problems are relatively idealized, gaining a precise
understanding of methods in these settings provides a firm foundation
for the analysis and improvement of RL algorithms in more complex
settings.  To provide some flavor for the quantitative trade-offs that
arise, in context of the \mbox{$d$-dimensional} linear quadratic
regulator, Tu and Recht~\cite{TuRec19} studied the LSTD algorithm, a
model-free method for policy evaluation, and proved that it has sample
complexity larger by factor $d$ than a model-based approach that
directly estimates the linear dynamics and then applies a robust
solver for the Ricatti equation.  As another example, in our own past
work~\cite{Wai19} on $\discount$-discounted MDPs with finite
state-action spaces, we have shown that the usual $Q$-learning suffers
from at least worst-case fourth-order scaling in the discount
complexity $1/(1-\discount)$, as opposed to the third-order scaling
that is achievable by empirical
$Q$-value-iteration~\cite{AzaMunKap13}.

In this paper, we revisit the classical problem of $Q$-learning in
MDPs with finite state-action spaces.  Our main contribution is to
introduce a simple variant of $Q$-learning based on an appropriate
form of variance reduction, and to prove that up to a logarithmic
factor in discount complexity, it achieves the minimax optimal sample
complexity~\cite{AzaMunKap13} for estimating $Q$-functions in
$\ell_\infty$-norm.

\paragraph{Related work and our contributions}

In this paper, we study $\discount$-discounted Markov decision
processes with finite state space $\StateSpace$ and action space
$\ActionSpace$.  Throughout, we adopt the shorthand $D \defn
|\StateSpace| \times |\ActionSpace|$ for the total number of
state-action pairs.  Our main focus is the performance of iterative
algorithms for estimating the optimal $Q$-function in the
$\ell_\infty$-norm, and our brief overview of the past literature is
accordingly targeted.  The $Q$-learning algorithm itself is
classical~\cite{WatDay92}, and there is a long line of work on its
analysis (e.g.,~\cite{Tsi94, JaaJorSin94, Sze97,KeaSin99,
  BerTsi96,EveMan03, Wai19CC}).  Moreover, a number of extensions to
$Q$-learning have been proposed over the years
(e.g.,~\cite{KeaSin99,Aza11,LatHut14,JinZhuBubJor18}).  For the
$Q$-learning algorithm itself, our own recent work~\cite{Wai19CC}
established sharp upper bounds on the number of samples required to
achieve an $\epsilon$-accurate estimate of the optimal $Q$-function in
\mbox{$\ell_\infty$-norm.}  Consider in particular, the synchronous or
generative setting of $Q$-learning, in which at each iteration, we
observe a new state drawn from the transition probability distribution
indexed by each state-action pair.  In this setting, the sample
complexity of an algorithm corresponds to the total number of
state-action samples required to achieve an error of $\epsilon$; to be
clear, in the generative setting, the sample complexity is a factor of
$D$ larger than the number of iterations, since each iteration
involves drawing $D$ samples.  Various algorithms can be compared
based on their sample complexity.  For ordinary $Q$-learning, the best
known upper bound on the sample complexity required to achieve
$\epsilon$-accuracy in the $\ell_\infty$-norm scales as $\order
\left(\frac{\rmax^2}{\epsilon^2} \frac{D \log(
  D/\delta)}{(1-\discount)^5} \right)$, as shown in the
paper~\cite{Wai19CC}.  Earlier work by Azar et al.~\cite{Aza11} had
introduced an extension of $Q$-learning known as the speedy
$Q$-learning algorithm, and shown that it has sample complexity
$\order \left(\frac{\rmax^2}{\epsilon^2} \frac{D \log(
  D/\delta)}{(1-\discount)^4} \right)$.  In another piece of earlier
work, Azar et al.~\cite{AzaMunKap13} studied the sample complexity of
model-based $Q$-value-iteration---that is, in which the transition
probability matrices are estimated using a collection of data, and
then we perform $Q$-value iteration using the fitted model.  Note that
this can be viewed as a fully batched method, since it uses all the
data at once to form the empirical Bellman operator. Under the same
assumptions as above, they proved that this batched form of
$Q$-value-iteration yields an $\epsilon$-accurate estimate with
probability at least $1-\delta$ using a total of $\order
\left(\frac{\rmax^2}{\epsilon^2} \frac{D \log(
  D/\delta)}{(1-\discount)^3} \right)$ samples.  Moreover, they proved
that this sample complexity is minimax optimal.  In a more recent line
of work, brought to our attention after initial posting of this work,
Sidford et al.~\cite{Sid18a,Sid18b} substantially strengthened the
results of Azar et al.~\cite{AzaMunKap13}, in particular by showing
that a mini-batched form of value iteration, together with a form of
variance reduction applied to the batch operators, is not only minimax
optimal in estimating the value functions, but also can be used to
return a policy whose value function is $\epsilon$-close to the true
value function.  This strengthening requires an algorithm that
carefully maintains certain monotonicity relations at each iterate,
along with some delicate analysis. In other earlier work, Lattimore
and Hutter~\cite{LatHut14}, working in the more challenging on-line
setting, studied an extension of the UCLR algorithm, and proved that
it achieves the optimal $1/(1-\discount)^3$ scaling in the discount
complexity parameter.  However, their sample complexity bound either
requires restrictions on the state transition matrices, or has
quadratic scaling in the number of states (as opposed to the optimal
linear scaling).  To the best of our knowledge, it remains unresolved
as to whether this minor gap is intrinsic to the method or an artifact
of the analysis.

With this past work in context, a natural question is whether there is
a simple extension of the standard $Q$-learning algorithm that is
minimax optimal.  The main contribution of this paper is to answer
this question in the affirmative, up to a logarithmic factor.  In
order to do so, we introduce an extension of $Q$-learning based on an
appropriate form of variance reduction.  To be clear, variance
reduction in stochastic approximation is a well-known idea, shown to
be especially fruitful in accelerating stochastic gradient methods for
optimization (e.g.,~\cite{SchRouBac17,ShaZha13, DefBacLac14,
  JohZha13}); in the context of reinforcement learning, it has also
been applied independently in the context of variance-reduced value
iteration~\cite{Sid18a,Sid18b} as well as policy
iteration~\cite{Du17}.

The form of variance-reduced $Q$-learning that we study, to be
specified in Section~\ref{SecMain}, is relatively simple to describe
and implement, and can be seen to be using the same variance-reduction
device as the SVRG algorithm in stochastic
optimization~\cite{JohZha13}.  Our main result is a sharp analysis of
this procedure, showing that it has minimax optimal sample
complexity~\cite{AzaMunKap13} up to a logarithmic factor in the
discount complexity $1/(1-\discount)$.  Analysis of variance-reduced
$Q$-learning requires techniques different from those used in
stochastic optimization, in particular building off the non-asymptotic
bounds for cone-contractive operators introduced in our past
work~\cite{Wai19CC}, as well as recent work~\cite{AzaMunKap13,
  LatHut14} in reinforcement learning that provides control on the
variance of the empirical Bellman operator and related quantities.

The remainder of this paper is organized as follows.  We begin in
Section~\ref{SecBackground} with basic background on Markov decision
processes and the $Q$-learning algorithm.  In Section~\ref{SecMain},
we introduce the variance-reduced $Q$-learning algorithm studied in
this paper, and state our main results (Theorem~\ref{ThmVarRed},
Corollary~\ref{CorSampleSize} and Proposition~\ref{PropMinimax}) on
its convergence guarantees.  Section~\ref{SecProof} is devoted to the
proof of our main results, with the proofs of some auxiliary results
provided in the appendix.

\paragraph{Notation:}  Throughout the paper, we use notation such as
$\unicon$, $\uniconprime$ etc. to denote universal constants that do
not depend on any parameters of the MDP, including the discount factor
$\discount$, size of state and action spaces and so on.  A warning to
the reader: the values of these universal constants may change from
line to line within an argument.


\section{Background}
\label{SecBackground}

We begin by providing some standard background on Markov decision
processes and the $Q$-learning algorithm, before discussing the
effective variance in $Q$-learning.  Our treatment is very brief; we
refer the reader to various books (e.g.,~\cite{Puterman05, SutBar18,
  Bertsekas_dyn1, BerTsi96, Sze09}) for more background on MDPs and
reinforcement learning.

\subsection{Markov decision processes and $Q$-functions}

In this paper, we study Markov decision process (MDP) with a finite
set of possible states $\StateSpace$, and a finite set of possible
actions $\ActionSpace$.  The states evolve dynamically in time, with
the evolution being influenced by the actions.  More precisely, we
define a collection of probability transition functions $\{
\Tfun{\state}{\action}{\cdot} \mid (\state, \action) \in \StateSpace
\times \ActionSpace \}$, indexed by state-action pairs $(\state,
\action)$.  When in state $\state$, performing an action $\action$
causes a transition to the next state drawn randomly from the
transition function $\Tfun{\state}{\action}{\cdot}$.  The next
ingredient of an MDP is a reward function $\reward$; it maps
state-action pairs to real numbers, so that $\reward(\state, \action)$
is the reward received upon executing action $\action$ while in state
$\state$.  A deterministic policy $\pol$ is a mapping from the state
space to the action space, so that action $\pol(\state)$ is taken when
in state $\state$.

The quality of a policy is measured by the expected sum of
discounted rewards over all state-action pairs in an infinite sample
path.  Of central interest to this paper is the
\emph{$Q$-value-function or state-action function} associated with a
given policy $\pol$.  For a given discount factor $\discount \in
(0,1)$, it is given by
\begin{align}
\Qpol(\state, \action) & \defn \Exs \left [ \sum^\infty_{\titer=0}
  \discount^\titer \reward(\state_\titer, \action_\titer) \mid
  \state_0 = \state, \action_0 = \action \right] \qquad \mbox{where
  $\action_\titer = \pol(\state_\titer)$ for all $\titer \geq 1$.}
\end{align}
That is, it measures the expected sum of discounted rewards,
conditioned on starting in state-action pair $(\state, \action)$, and
following the policy $\pol$ in  all subsequent transitions.

\subsection{Bellman operators and $Q$-learning}

Naturally, we would like to choose the policy $\pol$ so as to optimize
the values of the $Q$-function.  From the classical theory of finite
Markov decision processes~\cite{Puterman05, SutBar18, BerTsi96}, it is
known that there exists an optimal deterministic policy, and it can be
found by computing the unique fixed point of the Bellman operator.
The Bellman operator is a mapping from $\real^{|\StateSpace| \times
  |\ActionSpace|}$ to itself, whose $(\state, \action)$-entry is given
by
\begin{align}
\label{EqnPopBellman}
  \Bellman(\theta)(\state, \action) & \defn \reward(\state, \action) +
  \discount \Exs_{\xstate{}'} \max_{\action' \in \ActionSpace}
  \theta(\xstate{}', \action') \qquad \mbox{where $\xstate{}' \sim
    \Tfun{\state}{\action}{\cdot}$.}
\end{align}
It is well-known that $\Bellman$ is $\discount$-contractive with
respect to the $\ell_\infty$-norm
\begin{align}
  \|\theta\|_\infty \defn \max \limits_{(\state, \action)
    \in \StateSpace \times \ActionSpace} |\theta(\state,
  \action)|.
\end{align}
This property ensures the existence and uniqueness of a fixed point
$\thetastar$, and any optimal policy takes the form $\polopt(\state)
\in \arg \max \limits_{\action \in \ActionSpace} \thetastar(\state,
\action)$.

In the learning context, the transition dynamics
$\{\Tfun{\state}{\action}{\cdot}, \; (\state, \action) \in \StateSpace
\times \ActionSpace \}$ are unknown, so that it is not possible to
exactly evaluate the Bellman operator.  Instead, we assume some form
of access to a simulation engine that generates samples.  In this
paper, we study the \emph{synchronous or generative setting}, in which
at each time $\titer = 1, 2, \ldots$ and for \emph{each} state-action
pair $(\state, \action)$, we observe a sample $x_\titer(\state,
\action)$ drawn according to the transition function
$\Tfun{\state}{\action}{\cdot}$.  We note that guarantees for the
sychronous setting can be transferred to guarantees for the on-line
setting via notions of cover times of Markov chains; we refer the
reader to the papers~\cite{EveMan03,Aza11} for conversions of this
type.

The synchronous form of $Q$-learning algorithm generates a sequence of
iterates $\{\thetait{\titer} \}_{\titer \geq 1}$ according to the
recursion
\begin{align}
\label{EqnOrdinaryQ}
\thetait{\titer + 1} & = ( 1- \stepit{\titer}) \thetait{\titer} +
\stepit{\titer} \EmpBellmanIter{\titer}(\thetait{\titer}).
\end{align}
Here $\stepit{\titer} \in (0,1)$ is a stepsize to be chosen by the
user.  The operator $\EmpBellmanIter{\titer}$ is a mapping from
$\real^{|\StateSpace| \times |\ActionSpace|}$ to itself, and is known
as the \emph{empirical Bellman operator}: its $(\state,
\action)$-entry is given by
  \begin{align}
  \label{EqnEmpBellman}
  \EmpBellmanIter{\titer}(\theta)(\state, \action) = \reward(\state,
  \action) + \discount \max_{\action' \in \ActionSpace} \theta \big(
  \xstate{\titer}, \action' \big).
  \end{align}
  Here $\xstate{\titer} \in \real^{|\StateSpace| \times
    |\ActionSpace|}$ is a random matrix indexed by state-action pairs
  $(\state, \action)$; entry $\xstate{\titer}(\state, \action)$ is
  drawn according to the probability distribution
  $\Tfun{\state}{\action}{\cdot}$.  By construction, for any fixed
  $\theta$, we have $\Exs[ \EmpBellmanIter{\titer}(\theta)] =
  \Bellman(\theta)$, so that the empirical Bellman
  operator~\eqref{EqnEmpBellman} is an unbiased estimate of the
  population Bellman operator~\eqref{EqnPopBellman}.  Thus, we
  recognize $Q$-learning as a particular form of stochastic
  approximation.

For future reference, it is worth noting that
$\EmpBellmanIter{\titer}$ is also $\discount$-contractive with respect
to the $\ell_\infty$-norm; in particular, we have
  \begin{align}
    \label{EqnEmpContractive}
    \|\EmpBellmanIter{\titer}(\theta) -
    \EmpBellmanIter{\titer}(\theta')\|_\infty & \leq \discount \|
    \theta - \theta'\|_\infty
  \end{align}
  for any pair of $Q$-functions $\theta$ and $\theta'$, as can be
  verified by direct calculation.

\subsection{The effective variance in ordinary $Q$-learning}

As is well known from the theory of stochastic approximation, the
accuracy of iterative procedures like
$Q$-learning~\eqref{EqnOrdinaryQ} is partly controlled by the variance
of the updates.  In order to make this intuition clear, let us
introduce the error matrix $\DelIt{\titer} = \thetait{\titer} -
\thetastar$, and rewrite the $Q$-learning updates~\eqref{EqnOrdinaryQ}
in the recentered form
\begin{align}
  \label{EqnOrdinaryQError}
\DeltaIt{\titer+1} & = (1 - \stepit{\titer}) \DelIt{\titer} +
\stepit{\titer} \left \{ \EmpBellmanIter{\titer}(\thetastar +
\DeltaIt{\titer}) - \EmpBellmanIter{\titer}(\thetastar) \right\} +
\stepit{\titer} V_\titer.
\end{align}
Here $V_\titer \defn \EmpBellmanIter{\titer}(\thetastar) -
\Bellman(\thetastar)$ is a zero-mean random matrix, in which entry
$(\state, \action)$ has variance
\begin{align}
  \label{EqnBellmanVar}
\sigma^2(\thetastar)(\state, \action) & \defn \var
\left(\EmpBellmanIter{\titer}(\thetastar)(\state, \action) \right).
\end{align}
The matrix of variances $\sigma^2(\thetastar)$ controls the asymptotic
behavior of the algorithm, and it plays a central role in our non-asymptotic
analysis in the sequel.


\section{Variance-reduced $Q$-learning}
\label{SecMain}

In this section, we give a precise specification of the
variance-reduced $Q$-learning algorithm studied in this paper.  Before
doing so in Section~\ref{SecAlg}, we begin in Section~\ref{SecOracle}
with some intuition from an oracle form of variance reduction.  In
Section~\ref{SecTheory}, we state our main theoretical result
(Theorem~\ref{ThmVarRed}) on variance-reduced $Q$-learning, along with
a follow-up result (Proposition~\ref{PropMinimax}) that shows its
minimax optimality up to a logarithmic factor.


\subsection{$Q$-learning with oracle variance reduction}
\label{SecOracle}

We begin with a thought experiment about an algorithm that, while
neither implementable nor sensible---because it assumes access to the
quantity $\thetastar$ that we are trying to compute---nonetheless
provides helpful intuition. More precisely, suppose that we could
compute both an empirical Bellman update
$\EmpBellmanIter{\titer}(\thetastar)$ and the population\footnote{Of
  course, we have $\Bellman(\thetastar) = \thetastar$ by definition,
  but we write the update in this way to build a natural bridge to our
  form of variance-reduced $Q$-learning.} Bellman update
$\Bellman(\thetastar)$.  In this case, we could implement the
recentered ``algorithm''
\begin{subequations}
\begin{align}
  \label{EqnIdealized}
\thetait{\titer + 1} & = (1 - \stepit{\titer}) \thetait{\titer} +
\stepit{\titer} \left( \EmpBellmanIter{\titer}(\thetait{\titer}) -
\EmpBellmanIter{\titer}(\thetastar) + \Bellman(\thetastar) \right \}.
\end{align}
What is the effective variance of these updates?  Again defining the
error matrix $\DelIt{\titer} \defn \thetait{\titer} - \thetastar$, we
find that it evolves according to the recursion:
\begin{align}
  \label{EqnIdealizedError}
  \DelIt{\titer +1} & = (1 - \stepit{\titer}) \DelIt{\titer} +
  \stepit{\titer} \left \{ \EmpBellmanIter{\titer}(\thetastar +
  \DelIt{\titer}) - \EmpBellmanIter{\titer}(\thetastar) \right \}.
\end{align}
\end{subequations}
By construction, this is entirely analogous to the evolution of the
error matrix in ordinary $Q$-learning~\eqref{EqnOrdinaryQError}, but
without the additional additive noise term.  Moreover, from the
$\discount$-contractivity of the empirical Bellman
update~\eqref{EqnEmpContractive}, each of updates $\Delta \mapsto
\EmpBellmanIter{\titer}(\thetastar + \Delta) -
\EmpBellmanIter{\titer}(\thetastar)$, for $\titer = 1, 2, \ldots,$ is
$\discount$-contractive in $\ell_\infty$-norm.  Consequently, if we
were to run this idealized algorithm~\eqref{EqnIdealized} with a
constant step size, the error matrix $\DeltaIt{\titer}$ from the
idealized update~\eqref{EqnIdealizedError} would vanish at a geometric
rate.

While the idealized update is not implementable, it gives intuition
for the form of variance reduction that we study.  Given an algorithm
that converges to $\thetastar$, we can use one of its iterates
$\mythetabar$ as a proxy for $\thetastar$, and then recenter the
ordinary $Q$-learning updates by the quantity $-
\EmpBellmanIter{\titer}(\mythetabar) + \Bellman(\mythetabar)$.  Note
that even this recentering is not implementable, since we cannot
compute the population Bellman update $\Bellman(\mythetabar)$ exactly.
However, we can use a set of samples to generate an unbiased
approximation of it.  In a nutshell, this is the form of
variance-reduced $Q$-learning that we study.


\subsection{An implementable form of variance reduction} 
\label{SecAlg}

With this intuition in hand, let us now describe the form of
variance-reduced $Q$-learning that we study in this paper.  At the
core of the algorithm is a variance-reduced form of $Q$-learning,
which we describe Section~\ref{SecVarReduce}.  The algorithm itself
consists of a sequence of epochs, and we specify the form of each
epoch in Section~\ref{SecSingEpoch}. We combine these ingredients to
specify the overall algorithm in Section~\ref{SecOverall}.


\subsubsection{The basic variance-reduced update}
\label{SecVarReduce}

We begin by defining a sequence of operators $\{ \VARUP_\titer
\}_{\titer \geq 1}$ that define the variance-reduced $Q$-learning
algorithm.  Recall from our previous discussion that the method uses a
matrix $\mythetabar$ as a surrogate to $\thetastar$, and requires an
approximation to the Bellman update $\Bellman(\mythetabar)$.  In
particular, for a given integer $\N \geq 1$, the parameters of the
algorithm, we define the Monte Carlo approximation
\begin{align}
\BellmanMixN(\mythetabar) & \defn \frac{1}{\N} \sum_{i \in \SamN}
\EmpBellmanIter{i}(\mythetabar)
\end{align}
where $\SamN$ is a collection of $\N$ i.i.d. samples (i.e., matrices
with samples for each state-action pair $(\state, \action)$).  By
construction, the random matrix $\BellmanMixN(\mythetabar)$ is an
unbiased approximation of population Bellman update
$\Bellman(\mythetabar)$.  Each of its entries has variance
proportional to $1/\N$, so that we can control the approximation error
by a suitable choice of $\N$.

Given the pair $(\mythetabar, \BellmanMixN(\mythetabar))$ and a
stepsize parameter $\stepit{} \in (0,1)$, we define an operator
$\VARUP_\titer$ on $\real^{|\StateSpace| \times |\ActionSpace|}$ via
\begin{align}
  \label{EqnVarRedUpdates}
\theta \mapsto \VARUP_\titer(\theta; \stepit{}, \mythetabar,
\BellmanMixN) & \defn (1 - \stepit{}) \theta + \stepit{} \left \{
\EmpBellmanIter{\titer}(\thetait{}) -
\EmpBellmanIter{\titer}(\mythetabar) + \BellmanMixN(\mythetabar)
\right \}.
\end{align}
Here $\EmpBellmanIter{\titer}$ is a version of the empirical Bellman
operator constructed using a sample \emph{not} in $\SamN$.  Thus, the
random operators $\EmpBellmanIter{\titer}$ and $\BellmanMixN$ are
independent.  By construction, we have
\begin{align*}
\Exs \left[ \EmpBellmanIter{\titer}(\thetait{}) -
  \EmpBellmanIter{\titer}(\mythetabar) + \BellmanMixN(\mythetabar)
  \right] & = \Bellman(\thetait{})
\end{align*}
so that this update is unbiased as an estimate of the population
Bellman update.  As noted earlier, the device used to construct the
variance-reduced operator~\eqref{EqnVarRedUpdates} is the same as that
used in the SVRG algorithm for stochastic
optimization~\cite{JohZha13}.


\subsubsection{A single epoch}
\label{SecSingEpoch}

Having defined the basic variance-reduced
update~\eqref{EqnVarRedUpdates}, we now describe how to exploit in a
sequence of epochs.  The input to each epoch is a matrix
$\mythetabar$, corresponding to our current best guess of the optimal
$Q$-function $\thetastar$.  Epochs are parameterized by their length
$\EpLength$, corresponding to the number of iterations of the variance
reduced update, and a second integer $\N$, corresponding to the number
of samples used to compute the Monte Carlo approximation
$\BellmanMixN$.  We summarize the operation of an epoch in terms of the
following function $\texttt{RunEpoch}$:
\myframe{ {\bf{Function
      $\texttt{RunEpoch}(\mythetabar; \EpLength, \N)$}}
  \\ {\bf{Inputs:}}\\
  \begin{tabular}{ccccc}
  $(a)$ Epoch length $\EpLength$ && $(b)$
    Recentering matrix $\mythetabar$ && $(c)$ Recentering
    sample size $\N$
  \end{tabular}
  \begin{enumerate}
\item[(1)] Compute $\BellmanMixN(\mythetabar)  \defn \frac{1}{\N}
  \sum_{i=1}^\N \EmpBellmanIter{i}(\mythetabar)$.
\item[(2)] Initialize $\thetait{1} = \mythetabar$.
\item[(3)] For $\titer = 1, \ldots, \EpLength$, compute the
  variance-reduced update~\eqref{EqnVarRedUpdates}:
  \begin{align}
    \thetait{\titer +
    1} = \VARUP_\titer(\thetait{\titer}; \stepit{\titer}, \mythetabar,
  \BellmanMixN) \qquad \mbox{with stepsize $\stepit{\titer} = \frac{1}{1 +
      (1-\discount) \titer}$.}
  \end{align}
  \end{enumerate}
      {\bf{Output:}} Return $\thetait{\EpLength+1}$.  }
The choice of stepsize $\stepit{\titer} = \frac{1}{1 + (1 - \discount)
  \titer}$ is motivated by our previous work on ordinary
$Q$-learning~\cite{Wai19CC}, where we proved sharp non-asymptotic
bounds with this choice.  We use this same approach in analyzing the
behavior of the variance-reduced updates (Step (3) in
$\texttt{RunEpoch}$) within each epoch.  (It is worth noting that past
work~\cite{Sze97,EveMan03} shows that the stepsize choice
$\stepit{\titer} = 1/\titer$ leads to very poor behavior with ordinary
$Q$-learning---in particular, a convergence rate that is exponentially
slow in terms of the discount complexity parameter---and the same
statement would apply to our variance-reduced updates.)

\subsubsection{Overall algorithm}
\label{SecOverall}

We now have the necessary ingredients to specify the variance-reduced
$Q$-learning algorithm.  The overall algorithm is parameterized by
three choices: the total number of epochs $\NumEpoch \geq 1$ to be
run; the length $\EpLength$ of each epoch; and the sequence of
recentering samples $\{\Nit{\eiter}\}_{\eiter=1}^\NumEpoch$ used in
the $\NumEpoch$ epochs.  Each epoch is based on a single call to the
function $\texttt{RunEpoch}$.  Over all the epochs, the total number
of matrix samples used in any run of the algorithm is given by
$\EpLength \NumEpoch + \sum_{\eiter = 1}^\NumEpoch \Nit{\eiter}$.
Given any choice of the triple $(\NumEpoch, \EpLength,
\{\Nit{\eiter}\}_{\eiter=1}^\NumEpoch)$, the overall algorithm takes
the following form: \\

\vspace*{.1in}
\myframe{
{\bf{Algorithm: Variance-reduced $Q$-learning}} \\
{\bf{Inputs:}}
\begin{tabular}{cccc}
$(a)$ Number of epochs $\NumEpoch$ && $(b)$ Epoch length $\EpLength$
  & $(c)$ Recentering sizes
  $\{\Nit{\eiter}\}_{\eiter=1}^\NumEpoch$
\end{tabular}
\begin{enumerate}
\item[(1)] Initialize $\mythetabar_0 = 0$.
\item[(2)] For epochs $\eiter = 1, \ldots, \NumEpoch$: $\qquad
  \mythetabar_{\eiter} = \texttt{RunEpoch}(\mythetabar_{\eiter-1};
  \EpLength, \Nit{\eiter})$.
\end{enumerate}
{\bf{Output:}}   Return $Q$-function estimate $\mythetabar_\NumEpoch$
}

For a given tolerance parameter $\delta \in (0,1)$, we we choose the
epoch length $\EpLength$ and recentering sizes
$\{\Nit{\eiter}\}_{\eiter=1}^\NumEpoch$ so as to ensure that our final
guarantees hold with probability at least $1-\delta$.  The dependence
on the failure probability $\delta$ scales as $\log(1/\delta)$.  For
the purposes of our analysis, we choose these parameters in the
following way:
\begin{subequations}
  \label{EqnParChoices}
  \begin{align}
    \label{EqnEpochLength}
    \mbox{{\bf{Epoch length}}} \qquad \qquad \EpLength & = \unicon_1
    \; \frac{\log \left(\frac{8 \NumEpoch \Dim}{(1-\discount) \delta}
      \right) }{(1-\discount)^3},
\end{align}
and in epochs $\epochit = 1, \ldots, \NumEpoch$, we use the
\begin{align}
  \label{EqnRecSamples}
    \mbox{{\bf{Recentering sample sizes}}} \qquad \qquad
    \Nit{\epochit} & = \unicon_2 \; 4^\epochit \frac{\log(8 \NumEpoch
      \Dim/\delta)}{(1-\discount)^2}.
\end{align}
\end{subequations}

A few comments about these choices are in order.  First, our choice of
the epoch length~\eqref{EqnEpochLength} serves to make the error
decrease by a factor of $1/2$ in each epoch.  A larger choice
$\EpLength$ is not helpful (and in fact, wastes samples), since the
effective noise in variance-reduced $Q$-learning includes an
additional bias term that persists independently of the number of
iterations within the epoch.  On the other hand, note that the number
of samples $\Nit{\eiter}$ used in epoch $\eiter$ follows the geometric
progression $4^\eiter$ as a function of the epoch number.  This
increase is needed in order to ensure that the bias introduced in our
estimate of the Bellman operator
$\BellmanMixN(\mythetabar_{\eiter-1})$ decreases geometrically as a
function of $\eiter$.  Our particular choice of the factor $4$ in the
geometric progression was for concreteness, and is not essential; as
shown in Figure~\ref{FigEpochErr}(b), the algorithm has qualitatively
similarly convergence behavior for other choices of this parameter.


\subsection{Theoretical guarantees}
\label{SecTheory}

In this section, we state our main theoretical guarantees on
variance-reduced $Q$-learning, beginning with its geometric
convergence rate as a function of the epoch number
(Theorem~\ref{ThmVarRed}), followed by an upper bound on the total
number of samples used (Corollary~\ref{CorSampleSize}).

\subsection{Geometric convergence over epochs}

Our main result guarantees that variance-reduced $Q$-learning exhibits
geometric convergence over the epochs with high probability.  More
precisely, we have:
\begin{theorem}
\label{ThmVarRed}
Given a $\discount$-discounted MDP with optimal $Q$-function
$\thetastar$ and a given error probability $\delta \in (0,1)$, suppose
that we run variance-reduced $Q$-learning from $\thetabarit{0} = 0$
for $\NumEpoch$ epochs using parameters $\EpLength$ and
$\{\Nit{\epochit}\}_{\epochit \geq 1}$ chosen according to the
criteria~\eqref{EqnParChoices}.  Then we have
\begin{align}
  \label{EqnVarRed}
  \|\thetabarit{\NumEpoch} - \thetastar\|_\infty \leq
  \frac{\|\sigma(\thetastar)\|_\infty + \|\thetastar\|_\infty
    (1-\discount) }{2^\NumEpoch} \qquad \mbox{with probability at
    least $1 - \delta$.}
\end{align}
\end{theorem}
An immediate consequence of Theorem~\ref{ThmVarRed} is that for any
$\epsilon > 0$, running the algorithm with the number of epochs
\begin{align}
  \label{EqnDefnToTepoch}
  \NumEpoch(\epsilon; \thetastar) & \defn \left \lceil \log_2
  \left(\frac{\|\sigma(\thetastar)\|_\infty + \|\thetastar\|_\infty
    (1-\discount)}{ \epsilon} \right )\right \rceil
\end{align}
yields an output that is $\epsilon$-accurate in $\ell_\infty$-norm,
with probability at least $1 - \delta$.

\subsubsection{Illustrations of qualitative behavior}

\begin{figure}[h]
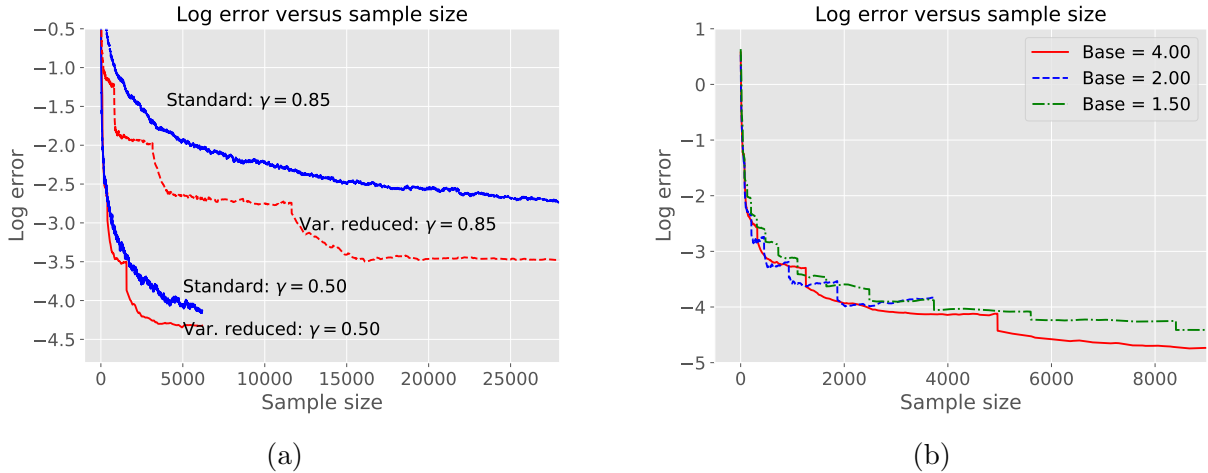

  \begin{center}
    \begin{tabular}{ccc}
    \widgraph{0.47\textwidth}{\figdir/fig_epoch_convergence} &&
    \widgraph{0.47\textwidth}{\figdir/fig_epoch_params} \\
      (a) && (b)
    \end{tabular}
    \caption{(a) Comparison of the convergence behavior of
      variance-reduced $Q$-learning and ordinary $Q$-learning with
      rescaled linear stepsize.  For each algorithm, the figure plots
      the \mbox{log $\ell_\infty$-error} \mbox{$\|\theta_n -
        \thetastar\|_\infty$} versus the number of samples $n$ for two
      different values of the discount factor \mbox{$\discount \in
        \{0.50, 0.85 \}$.}  As predicted by our theory, the gains from
      variance-reduction become significant as $\discount$ increases
      towards $1$. (b) Behavior of variance-reduced $Q$-learning for
      different choices of the epoch reduction factor (base).  In our
      proof, we established the result for the choice $2.0$, but other
      choices are also valid, modulo slightly different choices of the
      parameters $\EpLength$ and $\{\Nit{\eiter} \}_{\eiter \geq 1}$.}
    \label{FigEpochErr}
  \end{center}
\end{figure}

In Figure~\ref{FigEpochErr}, we provide some plots that illustrate the
qualitative behavior of variance-reduced $Q$-learning.  In panel (a),
we plot the log \mbox{$\ell_\infty$-error} versus the number of
samples for both VR-Q-learning (red dashed curves), and ordinary
$Q$-learning (blue solid curves).  Due to the epoch structure of
VR-Q-learning, note how the error decreases in distinct
quanta.\footnote{We have interpolated the error so as to avoid sharp
  jumps while conveying the qualitative behavior.}  For small values
of the discount factor $\discount$, the convergence rate of
VR-Q-learning is very similar to that of ordinary $Q$-learning.  On
the other hand, as $\discount$ increases towards $1$, we start to see
the benefits of variance reduction, as predicted by our theory.

In Theorem~\ref{ThmVarRed}, we proved that the algorithm converges at
a geometric rate, with contraction factor $1/2$ in terms of the number
of epochs.  The factor of $2$ is a consequence of the term $4^m$ in
our choice~\eqref{EqnRecSamples} of the recentering sample sizes.
More generally, by replacing the factor of $4^m$ with a term of the
form $(C^2)^m$ for any $C > 1$, we can prove geometric convergence
with contraction factor $1/C$.  Panel (b) illustrates the qualitative
effects of varying the choice of the base parameter $C$ on the
convergence behavior of the algorithm.


\subsubsection{Total number of samples used}

We now state a corollary that provides an explicit bound on the number
of samples required to return an $\epsilon$-accurate solution with
high probability, as a function of the instance $\thetastar$.  We then
specialize this result to the worst-case setting.   In stating this result,
we introduce the complexity parameter,
\begin{align*}
\Bou_0(\thetastar, \discount) \equiv \Bou_0 \defn
\|\sigma(\thetastar)\|_\infty + \|\thetastar\|_\infty (1-\discount),
\end{align*}
and recall that the number of epochs is given by $\NumEpoch = \left
\lceil \log_2 \left(\frac{\bou_0}{ \epsilon} \right ) \right \rceil$.

\begin{corollary}
  \label{CorSampleSize}
Consider a $\discount$-discounted MDP with optimal $Q$-function
$\thetastar$, a given error probability $\delta \in (0,1)$ and
$\ell_\infty$-error level $\epsilon > 0$.  Then there are universal
constants $\unicon, \uniconprime$ such that a total of
\begin{align}
 \label{EqnSampleSize}
  \Nsamp(\thetastar, \delta, \epsilon) & = \left \{ \unicon \:
  \frac{\log \left(\frac{8 \NumEpoch \Dim}{(1-\discount) \delta}
    \right) }{(1-\discount)^3} \log \left(\frac{\Bou_0}{\epsilon}
  \right) + \uniconprime \: \left(\frac{\Bou_0}{ \epsilon} \right )^2
  \frac{\log(8 \NumEpoch \Dim/\delta)}{(1-\discount)^2} \right \}
\end{align}
matrix samples in the generative model is sufficient to obtain an
$\epsilon$-accurate estimate with probability at least $1-\delta$.
\end{corollary}
\noindent See Section~\ref{SecSampleSize} for the proof of this claim. \\

Note that the bound~\eqref{EqnSampleSize} depends on the instance via
the quantities $\|\sigma(\thetastar)\|$ and $\|\thetastar\|_\infty$,
both of which can vary substantially as a function of $\thetastar$.
In order to obtain worst-case bounds, we study the the class
$\MDP(\discount, \rmax)$ of all optimal $Q$-functions that can be
obtained from a $\discount$-discounted MDP with an $\rmax$-uniformly
bounded reward function.  (The reward function is $\rmax$-uniformly
bounded means that $\max \limits_{(\state, \action) \in \StateSpace
  \times \ActionSpace} |\reward(\state, \action)| \leq \rmax$.)  Over
this class, it can be shown that
\begin{align*}
\sup_{\thetastar \in \MDP(\discount, \rmax)} \Bou_0(\thetastar,
\discount) & \leq \rmax \left \{ \frac{2}{1-\discount} + 1 \right \}
\; \leq \; \frac{4 \rmax}{1-\discount}.
\end{align*}
Applying this upper bound to equation~\eqref{EqnSampleSize} and
simplifying, we find the uniform upper bound
\begin{align}
\label{EqnAnnoying}
\sup_{\thetastar \in \MDP(\discount, \rmax)} \Nsamp(\thetastar,
\delta, \epsilon) & \leq \left \lceil \unicon \left(
\frac{\rmax^2}{\epsilon^2} \right) \frac{\log \left(\frac{
    \Dim}{(1-\discount) \delta} \right) \log
  \left(\frac{1}{(1-\discount)\epsilon} \right)}{(1-\discount)^4}
\right \rceil.
\end{align}
This worst-case upper bound improves upon the best known bounds for
ordinary $Q$-learning~\cite{Wai19CC}, but does not match the cubic
scaling in $1/(1-\discount)$ of the minimax optimal sample
complexity~\cite{AzaMunKap13}.  In the following section, we show that
a slightly refined analysis of our variance-reduced updates allows us
to achieve the minimax optimal sample complexity.

\subsubsection{Refining the worst-case guarantee}

Suppose that we run the algorithm analyzed in Theorem~\ref{ThmVarRed}
with $\epsilon = \frac{\rmax}{\sqrt{1 - \discount}}$.  From the
bound~\eqref{EqnAnnoying}, doing so requires
$\frac{1}{(1-\discount)^3}$ samples, up to the logarithmic factor
corrections.  Moreover, the output of this procedure---call it
$\thetabarit{0}$---then satisfies the bound $\|\thetabarit{0} -
\thetastar\|_\infty \leq \frac{\rmax}{\sqrt{1-\discount}}$ with high
probability.  We claim that running the variance-reduced updates from
this initialization for a further logarithmic number of steps yields
an algorithm with minimax-optimal sample complexity.

\begin{proposition}[Minimax optimality]
\label{PropMinimax}
Consider a $\discount$-discounted MDP with optimal $Q$-function
$\thetastar$, a given error probability $\delta \in (0,1)$, and a
given error tolerance.  Then running variance-reduced $Q$-learning
from an initial point $\thetabarit{0}$ such that $\|\thetabarit{0} -
\thetastar\|_\infty \leq \frac{\rmax}{\sqrt{1-\discount}}$ for a total
of $\NumEpoch = \unicon \log(\frac{\rmax}{(1-\discount) \epsilon})$
epochs using $\EpLength$ and $\{\Nit{\epochit}\}_{\epochit \geq 1}$
chosen according to the criteria~\eqref{EqnParChoices}, yields a
solution $\thetabarit{\NumEpoch}$ such that
\begin{align}
\|\thetabarit{\NumEpoch} - \thetastar\|_\infty & \leq \epsilon
\quad \mbox{with probability at least $1 - \delta$.}
\end{align}
\end{proposition}

The total number of matrix samples, counting both the initial
iterations required to obtain the initialization $\thetabarit{0}$ and
all later iterations, used to obtain this $\epsilon$-accurate solution
is at most
\begin{align}
\Tmmax(\discount, \epsilon, \delta) & \defn \unicon \left (
\frac{\rmax^2}{\epsilon^2} \right) \frac{\log
  \left(\frac{\Dim}{(1-\discount) \delta} \right)
  \log\left(\frac{1}{(1-\discount) \epsilon} \right)}{(1-\discount)^3}.
\end{align}
In this way, we have recovered the worst-case optimal cubic scaling in
$1/(1-\discount)$, matching the lower bound due to Azar at
al.~\cite{AzaMunKap13}.


\section{Proofs} 
\label{SecProof}

We now turn to the proof of Theorem~\ref{ThmVarRed}, as well as
Corollary~\ref{CorSampleSize} and Proposition~\ref{PropMinimax}.  In
all cases, we defer proofs of various auxiliary lemmas to the
appendices.


\subsection{Proof of Theorem~\ref{ThmVarRed}}

We begin with the proof of Theorem~\ref{ThmVarRed} on the geometric convergence
of variance-reduced $Q$-learning over epochs.


\subsubsection{High-level roadmap}
\label{SecRoadMap}

At a high-level, we prove Theorem~\ref{ThmVarRed} via an inductive
argument.  To set up the induction, we say that epoch $\eiter$
terminates successfully---or is ``good'' for short---if its output
$\thetabarit{\eiter}$ satisfies the bound
\begin{align}
  \label{EqnGoodBound}
  \|\thetabarit{\eiter} - \thetastar\|_\infty & \leq
  \frac{\|\sigma(\thetastar)\|_\infty + \|\thetastar\|_\infty
    (1-\discount) }{2^\eiter}.
\end{align}
Our strategy is to show that, with the specified choices of
$\EpLength$ and $\{\Nit{\eiter}\}_{\eiter \geq 1}$, the
bounds~\eqref{EqnGoodBound} hold uniformly with probability at least
$1 -\delta$, and we do so via induction on $\eiter$.  The inductive
argument consists of two steps.
  
\paragraph{Base case $\eiter = 1$:}
Given the initialization $\thetabarit{0} = 0$, we prove that
$\thetabarit{1}$ satisfies the bound~\eqref{EqnGoodBound} with
probability at least $1-\frac{\delta}{\NumEpoch}$.


\paragraph{Inductive step:}

In this step, we suppose that the input $\thetabarit{\eiter}$ to epoch
$\eiter$ satisfies the bound~\eqref{EqnGoodBound}.  We then prove that
$\thetabarit{\eiter + 1}$ satisfies the bound~\eqref{EqnGoodBound}
with probability at least $1- \frac{\delta}{\NumEpoch}$. \\

\paragraph{Union bound:} Finally, by taking a union bound over all
$\NumEpoch$ epochs of the algorithm, we are guaranteed that the
claim~\eqref{EqnGoodBound} holds uniformly for all $\eiter = 1,
\ldots, \NumEpoch$, with probability at least $1-\delta$.  This
implies the claimed bound~\eqref{EqnVarRed} in the theorem statement.


\subsubsection{Proof of the base case}

For the given initialization $\thetabarit{0} = 0$, note that we have
$\EmpBellmanIter{\titer}(\thetabarit{0}) = \reward$ and
$\BellmanMixN(\thetabarit{0}) = \reward$.  Consequently, we have
$\EmpBellmanIter{\titer}(\thetabarit{0}) -
\BellmanMixN(\thetabarit{0}) = 0$, so that the variance-reduced
updates~\eqref{EqnVarRedUpdates} reduce to the case of ordinary
$Q$-learning with stepsize $\stepit{\titer} = \frac{1}{1 +
  (1-\discount) \titer}$.  It follows from analysis in our past
work~\cite{Wai19CC} that there is a universal constant $\unicon' > 0$
such that, after $\NumEpoch$ iterations, we have
\begin{align}
  \label{EqnPassione}
  \|\thetait{\EpLength+1} - \thetastar\|_\infty & \leq \frac{
    \|\thetastar \|_\infty }{(1 - \discount) \EpLength} + \unicon'
  \left \{ \frac{ \|\sigma(\thetastar)\|_\infty \sqrt{\log (2 D
      \NumEpoch \EpLength/\delta)} }{(1-\discount)^{3/2}\;
    \sqrt{\EpLength }} + \frac{\|\thetastar\|_\infty \log
    \left(\frac{2 e D \NumEpoch \EpLength}{\delta} \left(1 +
    (1-\discount) \EpLength \right) \right)}{ \;(1-\discount)^2 \EpLength }
  \right \}
\end{align}
with probability at least $1 - \frac{\delta}{\NumEpoch}$, where the
matrix of variances $\sigma^2(\thetastar) = \var
\left(\EmpBellmanIter{}(\thetastar) \right)$ was defined
previously~\eqref{EqnBellmanVar}.

Consequently, choosing $\EpLength =
\unicon \frac{\log \left( \frac{8 \NumEpoch D}{\delta \,
    (1-\discount)} \right)}{(1-\discount)^3}$ for a sufficiently large
constant $\unicon$ suffices to ensure that
\begin{align*}
  \|\thetait{\EpLength+1} - \thetastar\|_\infty & \leq \frac{1}{2} \left
  \{ \|\sigma(\thetastar)\|_\infty + \|\thetastar\|_\infty
  (1-\discount) \right \} \qquad \mbox{with probability at least $1
    - \frac{\delta}{\NumEpoch}$.}
\end{align*}
Since $\thetabarit{1} = \thetait{\EpLength+1}$ by definition, this bound
is equivalent to the claim~\eqref{EqnGoodBound} for the base case
$\eiter = 1$, as desired.

\subsubsection{Proof of inductive step}

For this step, we assume that the input $\thetabarit{\eiter}$ to epoch
$\eiter$ satisfies the bound
\begin{align}
  \label{EqnGoodInit}
  \|\thetabarit{\eiter} - \thetastar\|_\infty \leq
  \underbrace{\frac{\|\sigma(\thetastar)\|_\infty +
      \|\thetastar\|_\infty (1-\discount) }{2^\eiter}}_{ = \,:
    \Bou_{\eiter}}
\end{align}
and our goal is to show that $\| \thetabarit{\eiter+1} -
\thetastar\|_\infty \leq \frac{\Bou_\eiter}{2}$.  Recall that
$\thetabarit{\eiter+1}$ is equivalent to the output
$\thetait{\EpLength}$ of running $\EpLength$ rounds of
variance-reduced $Q$-learning from the initialization $\thetait{0} =
\thetabarit{\eiter}$, using the parameter $\N = \Nit{\eiter}$ for the
operator $\BellmanMixN$.  In this section, we prove that there is a
universal constant $\unicon > 0$ such that
\begin{align}
  \label{EqnHardClaim}
\|\thetait{\EpLength + 1} - \thetastar\|_\infty & \leq \unicon \Bou_\eiter \; \left
\{ \frac{1}{1 + (1-\discount) \EpLength} +
\frac{1}{1- \discount} \; \sqrt{\frac{\log( 8 \NumEpoch
    \Dim \EpLength /\delta)}{1 + (1-\discount) \EpLength}} +
\sqrt{4^\eiter \frac{\log (8 \NumEpoch \Dim/\delta)}{(1 - \discount)^2
    \; \Nit{\eiter}}} \right \}
\end{align}
with probability at least $1 - \frac{\delta}{\NumEpoch}$.  From this
bound, we see that the choices of $\EpLength$ and $\Nit{\eiter}$ given
in equation~\eqref{EqnParChoices}, with sufficiently large constants
of the pre-factors $c_1$ and $c_2$, are sufficient to ensure that
$\|\thetait{\EpLength} - \thetastar\|_\infty \leq
  \frac{\Bou_\eiter}{2}$ with probability at least $1 -
  \frac{\delta}{\NumEpoch}$, as desired.  Accordingly, the remainder
  of this section is devoted to proving the
  bound~\eqref{EqnHardClaim}.

  Throughout the remainder of this proof, we drop the subscript
  $\eiter$ as it can be implicitly understood; in particular, we use
$\N$, $\Bou$ and $\mythetabar$ as shorthands for
  $\Nit{\eiter}$,  $\Bou_\eiter$ and $\mythetabar_\eiter$, respectively.\\


\boldpara{Rewriting the update}
Recall the form of the variance-reduced $Q$-learning
updates~\eqref{EqnVarRedUpdates}.  We begin by re-writing these
updates in a form suitable for analysis using the general results on
stochastic approximation from Wainwright~\cite{Wai19CC}.  Define the
recentered operators
\begin{align}
  \label{EqnRecenteredOps}
\HopK(\theta) \defn \EmpBellmanIter{\titer}(\theta) -
\EmpBellmanIter{\titer}(\thetastar), \quad \mbox{and} \quad
\HopTilN(\theta) \defn \BellmanMixN(\theta) -
\BellmanMixN(\thetastar).
\end{align}
By recentering the updates~\eqref{EqnVarRedUpdates} around the optimal
$Q$-function $\thetastar$, we can write
\begin{align}
\thetait{\titer+1} - \thetastar & = (1 - \stepit{\titer}) \left(
\thetait{\titer} - \thetastar \right) + \stepit{\titer} \left \{
\HopK(\thetait{\titer}) + \wnoiseit{\titer} \right \},
\end{align}
where $\wnoiseit{\titer} \defn  - \HopK(\mythetabar) -
\Bellman(\thetastar) + \BellmanMixN(\mythetabar)$ is a random noise
sequence.  We use this noise sequence to define an auxiliary
stochastic process
\begin{align}
  \label{EqnDefnPathit}
  \pathit{\titer} & \defn (1 - \stepit{\titer-1}) \pathit{\titer-1} +
  \stepit{\titer-1} \wnoiseit{\titer-1}, \qquad \mbox{with
    initialization $\pathit{1} = 0$.}
\end{align}
Note that the operator $\Hop_\titer(\theta) =
\EmpBellmanIter{\titer}(\theta) - \EmpBellmanIter{\titer}(\thetastar)$
is monotonic with respect to the orthant ordering, and
$\discount$-contractive with respect to the $\ell_\infty$-norm.
Consequently, from past results, we have:
\begin{corollary}[Adapted from the paper~\cite{Wai19CC}]
 \label{CorQbound}
For all iterations $\titer = 1, 2, \ldots$, we have
\begin{align}
  \label{EqnQbound}
\|\thetait{\titer + 1} - \thetastar\|_\infty & \leq \frac{2}{1 +
  (1-\discount) \titer} \left \{ \|\thetait{1} - \thetastar \|_\infty
+ \sum_{\iter=1}^{\titer} \| \pathit{\iter} \|_\infty \right \} +
\|\pathit{\titer+1}\|_\infty,
\end{align}
\end{corollary}

In order to derive a concrete result based on this bound, we need to
obtain high-probability upper bounds on the terms
$\|\pathit{\iter}\|_\infty$.  We begin by decomposing the effective
noise into a sum of three terms that can be controlled nicely.
Recalling the definition~\eqref{EqnRecenteredOps} of $\HopK$ and
$\HopTilN$, we have
\begin{align*}
\wnoiseit{\titer} & = - \HopK(\mythetabar) - \Bellman(\thetastar) +
\BellmanMixN(\mythetabar) \; = \; - \HopK(\mythetabar) +
\HopTilN(\mythetabar) + \left \{ \BellmanMixN(\thetastar) -
\Bellman(\thetastar) \right \}.
\end{align*}
Thus, if we define another recentered operator $\Hop(\theta) =
\Bellman(\theta) - \Bellman(\thetastar)$, then we have
\begin{align}
\label{EqnNoiseDecomp}
\wnoiseit{\titer} & = \underbrace{\left \{ \Hop(\mythetabar) -
  \HopK(\mythetabar) \right \}}_{\VnoiseA} + \underbrace{\left\{
  \HopTilN(\mythetabar) - \Hop(\mythetabar) \right \}}_{\VnoiseB} +
\underbrace{\left \{ \BellmanMixN(\thetastar) - \Bellman(\thetastar)
  \right \}}_{\VnoiseC}.
\end{align}
From the linearity of the recursion~\eqref{EqnDefnPathit} and the fact
that $\VnoiseB$ and $\VnoiseC$ are independent of $\titer$, we can
write
\begin{align*}
\pathit{\titer} & = \VnoiseB + \VnoiseC + \PnoiseA,
\end{align*}
where the stochastic process $\{\PnoiseA \}_{\titer \geq 1}$ evolves
according to a recursion of the form~\eqref{EqnDefnPathit} with
$\wnoiseit{\titer}$ replaced by $\VnoiseA$.  Applying the
bound~\eqref{EqnQbound} at iteration $\EpLength$ and using the fact
that $\|\thetait{1} - \thetastar\|_\infty \leq \Bou$ by assumption, we
find that
\begin{multline}
  \label{EqnQboundNew}
  \|\thetait{\EpLength + 1} - \thetastar\|_\infty \leq \frac{2 \Bou}{1
    + (1-\discount) \EpLength} + + 3 \left \{
  \frac{\|\VnoiseB\|_\infty + \|\VnoiseC\|_\infty}{1- \discount}
  \right \} + \left \{ \frac{ 2 \sum_{\iter=1}^{\EpLength} \|
    \pathit{\iter}'\|_\infty}{1 + (1-\discount) \EpLength} +
  \|\pathit{\EpLength + 1}'\|_\infty \right \}.
\end{multline}
Thus, it remains to bound the two terms on the right-hand side,
involving the noise terms $\VnoiseB$ and $\VnoiseC$, as well as the
stochastic process $\{\pathit{\titer}'\}_{\titer \geq 1}$.


\subsubsection{Bounding the recentering terms}

We begin by bounding the noise terms $\VnoiseB$ and $\VnoiseC$, which
arise from differences between the population Bellman operator
$\Bellman$ and the randomized approximation $\BellmanMixN$ used to
recenter the iterates throughout the given epoch.  Note that both
$\VnoiseB$ and $\VnoiseC$ are zero mean random variables, formed of
sums of $\N$ i.i.d. terms, so that we can control their magnitudes by
increasing $\N$.  The following lemma makes this intuition precise:

\begin{lemma}[High probability bounds on recentering terms]
  \label{LemBiasBounds}
Fix an arbitrary $\delta \in (0,1)$.  
  \begin{enumerate}
  \item[(a)]  If $\|\mythetabar - \thetastar\|_\infty \leq \Bou$, then
    \begin{subequations}
      \label{EqnBiasBounds}
      \begin{align}
        \label{EqnBouB}
\|\VnoiseB \|_\infty \leq 4 \Bou \sqrt{\frac{\log(8 \NumEpoch \Dim
    /\delta)}{\N}} \qquad \mbox{with prob. at least $1 -
  \frac{\delta}{3 \NumEpoch}$.}
\end{align}
\item[(b)] There is a universal constant $\unicon$ such that
  \begin{align}
    \label{EqnBouC}
    \|\VnoiseC\|_\infty & \leq \unicon \left \{
    \|\sigma(\thetastar)\|_\infty + \|\thetastar\|_\infty
    (1-\discount) \right \}  \: \sqrt{\frac{\log (8 \NumEpoch
        \Dim/\delta)}{\N}} \qquad \mbox{with prob. at least $1 -
      \frac{\delta}{3 \NumEpoch}$.}
\end{align}
    \end{subequations}
  \end{enumerate}
\end{lemma}
\noindent The proof is a relatively straightforward application of
Hoeffding's inequality (for the bound~\eqref{EqnBouB}) and Bernstein's
inequality (for the bound~\eqref{EqnBouC}). See
Appendix~\ref{SecProofLemBiasBounds} for the details.


\subsubsection{Bounding the process $\{\pathitprime{\titer}\}_{\titer \geq 1}$}

Our next step is to control the terms in the
bound~\eqref{EqnQboundNew} that depend on the stochastic process
$\{\pathitprime{\titer} \}_{\titer \geq 1}$.

\begin{lemma}[High probability bound on noise]
  \label{LemHighProbNoise}
There is a universal constant $\unicon > 0$ such that for any $\delta
\in (0,1)$
\begin{align}
  \label{EqnHighProbNoise}
    \left \{ \frac{ 2 \sum_{\iter=1}^\EpLength \| \pathit{\ell}'
      \|_\infty}{1 + (1-\discount) \EpLength} + \|\pathit{\EpLength
      +1}'\|_\infty \right \} & \leq \frac{\unicon \, \Bou}{1-
      \discount} \; \sqrt{\frac{2\log( 8 \NumEpoch \Dim \EpLength
        /\delta)}{1 + (1-\discount) \EpLength}}
\end{align}
with probability at least $1 - \frac{\delta}{3 \NumEpoch}$.
\end{lemma}
\noindent The proof of this lemma is more involved, in particular
involving an inductive argument to control the MGF of the process
$\{\pathitprime{\titer}\}_{\titer \geq 1}$.  See
Appendix~\ref{SecProofLemHighProbNoise} for the details.


\subsubsection{Putting together the pieces}

We now put together the pieces, in particular using the
bounds~\eqref{EqnBiasBounds} and~\eqref{EqnHighProbNoise} to control
the terms in the inequality~\eqref{EqnQboundNew}, with the ultimate
goal of proving the claim~\eqref{EqnHardClaim}.  Doing so yields that
there are universal constants such that
\begin{multline*}
\frac{\|\thetait{\EpLength + 1} - \thetastar\|_\infty}{\Bou} \leq
\frac{2}{1 + (1-\discount) \EpLength} 
  + \uniconprime \left \{ 1 + \frac{\|\sigma(\thetastar)\|_\infty +
    \|\thetastar\|_\infty (1-\discount) }{\Bou} \right \}
  \sqrt{\frac{\log (8 \NumEpoch \Dim/\delta)}{(1 - \discount)^2 \;
      \N}} \\
+ \frac{\unicon}{1- \discount} \; \sqrt{\frac{\log( 8 \NumEpoch \Dim
    \EpLength /\delta)}{1 + (1-\discount) \EpLength}}.
\end{multline*}
By union bound over the three different bounds that we have applied
(each holding with probability at least $1 - \frac{\delta}{3
  \NumEpoch}$), the overall bound holds with probability at least $1 -
\frac{\delta}{\NumEpoch}$.  Recalling that \mbox{$\Bou =
  \frac{\|\sigma(\thetastar)\|_\infty + \|\thetastar\|_\infty
    (1-\discount)}{2^\eiter}$,} we have
\begin{align*}
\left \{ 1 + \frac{\|\sigma(\thetastar)\|_\infty +
  \|\thetastar\|_\infty (1-\discount) }{\Bou} \right \}
\sqrt{\frac{\log (8 \NumEpoch \Dim/\delta)}{(1 - \discount)^2 \; \N}}
& \leq \unicon'' \sqrt{\frac{4^\eiter \log (8 \NumEpoch
    \Dim/\delta)}{(1 - \discount)^2 \: \N}}
\end{align*}
Putting together the pieces yields that, with probability at least $1
- \frac{\delta}{\NumEpoch}$, we have
\begin{align*}
 \frac{ \|\thetait{\EpLength + 1} - \thetastar\|_\infty}{\Bou} & \leq
 \unicon \left \{ \frac{1}{1 + (1-\discount) \EpLength} +
 \sqrt{\frac{4^\eiter \log (8 \NumEpoch \Dim/\delta)}{(1 -
     \discount)^2 \; \N}} +  \frac{1}{1- \discount} \;
 \sqrt{\frac{\log( 8 \NumEpoch \Dim \EpLength /\delta)}{1 +
     (1-\discount) \EpLength}} \right \}
\end{align*}
for some universal constant $\unicon$.  Thus, we have proved the
desired claim~\eqref{EqnHardClaim}, which completes the proof
of Theorem~\ref{ThmVarRed}.


\subsection{Proof of Corollary~\ref{CorSampleSize}}
\label{SecSampleSize}

Now let us prove the bound on the total number of samples used, as
stated in Corollary~\ref{CorSampleSize}.  Recall that we use
$\EpLength$ samples for the $Q$-learning updates within each epoch,
and $\Nit{\eiter}$ samples to compute the recentering operator
$\BellmanMix_{\Nit{\eiter}}$ in epoch $\eiter$.  Defining $\Bou_0
\defn \|\sigma(\thetastar)\|_\infty + \|\thetastar\|_\infty
(1-\discount)$, the total number of samples is used
\begin{align*}
\NumEpoch \EpLength + \sum_{\epochit=1}^\NumEpoch \Nit{\eiter} & \leq
\NumEpoch \EpLength + \unicon \; 4^{\NumEpoch} \frac{\log(8
  \NumEpoch \Dim/\delta)}{(1-\discount)^2} \\
& = \uniconprime \; \NumEpoch \frac{\log \left(\frac{8 \NumEpoch
    \Dim}{(1-\discount) \delta} \right) }{(1-\discount)^3} \log
\left(\frac{\BouZero}{\epsilon} \right) + \unicon \;
\left(\frac{\BouZero}{\epsilon} \right)^2 \; \frac{\log(8 \NumEpoch
  \Dim/\delta)}{(1-\discount)^2},
\end{align*}
as claimed in equation~\eqref{EqnSampleSize}.


\subsection{Proof of Proposition~\ref{PropMinimax}}
\label{SecMinimax}

We now turn to the proof of Proposition~\ref{PropMinimax}, which
applies a more refined analysis to guarantee the worst-case optimal
sample complexity.  At a high level, our proof is based on showing
that under the stated conditions, the epoch iterates
$\{\thetabarit{\epochit} \}_{\epochit=1}^\NumEpoch$ satisfy
\begin{align}
\|\thetabarit{\epochit} - \thetastar\|_\infty & \leq
\frac{1}{2^\epochit} \frac{\rmax}{\sqrt{1-\discount}} \qquad \mbox{for
  all $\epochit = 1, \ldots, \NumEpoch$}
\end{align}
with probability at least $1 - \delta$.

We follow the same inductive argument as before.  The base case is
trivial, so that it only remains to establish the inductive step. In
the epoch that moves from $\thetabarit{\epochit}$ to
$\thetabarit{\epochit + 1}$, variance-reduced $Q$-learning uses
$\thetabar \equiv \thetabarit{\epochit}$ in its recentering process.
By the induction hypothesis, we have
\begin{align}
  \label{EqnBaseAssumption}
\|\thetabar - \thetastar\|_\infty \leq \bou_\epochit \defn
\frac{1}{2^\epochit} \frac{\rmax}{\sqrt{1-\discount}}.
\end{align}
In this case, our analysis of the epoch is based on the two operators
\begin{subequations}
  \begin{align}
 \label{EqnHopJ}
\HopJEmpIter{\titer}(\theta) \defn \EmpBellmanIter{\titer}(\theta) -
\EmpBellmanIter{\titer}(\thetabar) + \BellmanMixN(\thetabar), \quad
\mbox{and} \quad \HopJ(\theta) \defn \Bellman(\theta) -
\Bellman(\thetabar) + \BellmanMixN(\thetabar).
\end{align}
Note that the variance-reduced $Q$-learning updates can be written as
\begin{align}
  \label{EqnHopJQvar}
\thetait{\titer+1} & = (1 - \stepit{\titer}) \thetait{\titer} +
\stepit{\titer} \HopJEmpIter{\titer}(\thetait{\titer}).
\end{align}
\end{subequations}
Moreover, note that $\HopJ$ is $\discount$-contractive, it has a
unique fixed point, which we denote by $\thetahat$.  Since
$\HopJ(\theta) = \Exs[\HopJEmpIter{\titer}(\theta)]$ by construction,
it is natural to analyze the convergence of $\thetait{\titer}$ to
$\thetahat$.

From the initialization $\thetabar$, suppose that we run the
variance-reduced updates with recentering point $\thetabar$ for
$\EpLength$ steps, thereby obtaining the estimate
$\thetait{\EpLength+1}$.  Given a bound on $\|\thetait{\EpLength+1} -
\thetahat\|_\infty$, we can then bound the distance to $\thetastar$
via the triangle inequality---viz.
\begin{align}
\label{EqnTriangle}
\|\thetait{\EpLength+1} - \thetastar\|_\infty & \leq
\|\thetait{\EpLength+1} - \thetahat\|_\infty + \|\thetahat -
\thetastar\|_\infty.
\end{align}
With this decomposition, our proof of Proposition~\ref{PropMinimax}
hinges on the following two lemmas.  The first lemma bounds the error
of $\thetait{\EpLength + 1}$ as an estimate of the fixed point
$\thetahat$:
\begin{lemma}
\label{LemSharpEpoch}
After $\EpLength = \unicon_1 \frac{\log \left(\frac{8 \NumEpoch
    \Dim}{(1-\discount) \delta} \right)}{(1-\discount)^3}$ iterations,
we are guaranteed that
\begin{align*}
  \|\thetait{\EpLength + 1} - \thetahat\|_\infty & \leq
  \frac{\bou_\epochit}{4} + \frac{1}{4} \|\thetahat -
  \thetastar\|_\infty \quad \mbox{with probability at least $1 -
    \frac{\delta}{2 \NumEpoch}$.}
\end{align*}
\end{lemma}
\noindent See Appendix~\ref{AppProofLemSharpEpoch} for the proof of
this claim. \\

\noindent
Note that $\thetahat$ is a fixed point of the operator $\HopJ$ from
equation~\eqref{EqnHopJ}, which can be seen as a perturbed version of
the original Bellman operator $\Bellman$, for which $\thetastar$ is the
fixed point.  Our second lemma uses this fact to bound the difference
$\thetahat - \thetastar$:
\begin{lemma}
  \label{LemSelfBounding}
  Given a sample size $\Nit{\eiter} = \unicon_2 4^\eiter \frac{\log
    (\NumEpoch \Dim/\delta)}{(1-\discount)^2}$, we have
  \begin{align}
    \|\thetahat - \thetastar\|_\infty & \leq \frac{\bou_\epochit}{5}
    \quad \mbox{with probability at least $1 - \frac{\delta}{2
        \NumEpoch}$.}
  \end{align}
\end{lemma}
\noindent See Appendix~\ref{AppProofLemSelfBounding} for the proof of
this claim. \\

Completing the inductive step is straightforward given these two
lemmas.  Combining with our earlier bound\label{EqnTriangle}, we have
\begin{align*}
\|\thetait{\EpLength+1} - \thetastar\|_\infty \; \stackrel{(i)}{\leq}
\; \left \{ \frac{\bou_\epochit}{4} + \frac{1}{4} \|\thetahat -
\thetastar\|_\infty \right \} + \|\thetahat - \thetastar \|_\infty & =
\frac{\bou_\epochit}{4} + \frac{5}{4} \|\thetahat -
\thetastar\|_\infty \\
& \stackrel{(ii)}{\leq} \frac{\bou_\epochit}{2},
\end{align*}
where steps (i) and (ii) follow from Lemmas~\ref{LemSharpEpoch}
and~\ref{LemSelfBounding}, respectively.  This sequence of
inequalities holds with probability at least $1 -
\frac{\delta}{\NumEpoch}$ as claimed.



\section{Discussion}

In this paper, we have proposed a variance-reduced form of
$Q$-learning, and shown that it has sample complexity that achieves
the minimax optimal sample complexity, up to a logarithmic factor in
the discount complexity $1/(1-\discount)$.  Although our result can be
summarized succinctly in this way, in fact, our analysis is instance
specific, and we have proved bounds that depend on the optimal
$Q$-function $\thetastar$ via its supremum norm
$\|\thetastar\|_\infty$, and the variance $\sigma^2(\thetastar)$ of
the associated empirical Bellman update.  Although the analysis of
this paper focuses purely on the tabular setting, the variance-reduced
$Q$-learning algorithm itself can be applied in more generality.  It
would be interesting to explore the uses of this algorithm in more
general settings.


\subsection*{Acknowledgements}
This work was partially supported by Office of Naval Research Grant
ONR-N00014-18-1-2640 and National Science Foundation Grant
NSF-DMS-1612948.  We thank A. Pananjady and K. Khamaru for careful
reading and comments on an earlier draft.


\appendix


\section{Auxiliary lemmas for Theorem~\ref{ThmVarRed}}
\label{AppProofThmVarRed}

In this appendix, we collect the proofs of Lemmas~\ref{LemBiasBounds}
and~\ref{LemHighProbNoise}, used in the proof of
Theorem~\ref{ThmVarRed}.


\subsection{Proof of Lemma~\ref{LemBiasBounds}}
\label{SecProofLemBiasBounds}

The lemma consists of the two separate bounds~\eqref{EqnBouB}
and~\eqref{EqnBouC}, and we split our proof accordingly.

\paragraph{Proof of bound~\eqref{EqnBouB}:}

By definition, the random operator $\HopTilN(\mythetabar) =
\frac{1}{\N} \sum_{i=1}^\N \big \{\EmpBellmanIter{i}(\mythetabar) -
\EmpBellmanIter{i}(\thetastar) \big \}$ is the sum of $\N$
i.i.d. terms.  Each random operator $\EmpBellmanIter{i}$ is
$\discount$-contractive, so that for each state-action pair $(\state,
\action)$, we have
\begin{align*}
\left |\EmpBellmanIter{i}(\mythetabar)(\state, \action) -
\EmpBellmanIter{i}(\thetastar)(\state, \action) \right| & \leq
\discount \|\mythetabar - \thetastar\|_\infty \; \leq \; \Bou.
\end{align*}
Consequently, each entry of the noise matrix $\VnoiseB$ is zero-mean,
and the i.i.d. sum of $\N$ random variables bounded in absolute value
by $\Bou$.  Therefore, the claim follows from Hoeffding's inequality
for bounded random variables~\cite{Wai19}.


\paragraph{Proof of bound~\eqref{EqnBouC}:}

Note that $\BellmanMixN(\thetastar) -\Bellman(\thetastar)$ is a sum of
$\N$ i.i.d. terms, each bounded in absolute value by
$\|\thetastar\|_\infty$, and with variance matrix
$\sigma^2(\thetastar)$, as was previously defined in
equation~\eqref{EqnBellmanVar}.  Consequently, by a combination of the
union bound (over state-action pairs) and Bernstein's inequality,
there is a universal constant $\unicon$ such that, with probability at
least $1 - \frac{\delta}{3 \NumEpoch}$, we have
\begin{align*}
\|\BellmanMixN(\thetastar) - \Bellman(\thetastar)\|_\infty & \leq c
\left \{ \|\sigma(\thetastar)\|_\infty \sqrt{ \frac{\log(8 \NumEpoch
    \Dim/\delta)}{ \N}} + \frac{\|\thetastar\|_\infty \log(8 \NumEpoch
  \Dim/\delta)}{\N} \right \} \\
& \leq \unicon \sqrt{ \frac{\log(8 \NumEpoch \Dim/\delta)}{ \N}} \left
\{ \|\sigma(\thetastar)\|_\infty + \|\thetastar\|_\infty \sqrt{
  \frac{\log(8 \NumEpoch \Dim/\delta)}{ \N}} \right \} \\
& \leq \unicon \sqrt{ \frac{\log(8 \NumEpoch \Dim/\delta)}{ \N}} \left
\{ \|\sigma(\thetastar)\|_\infty + \|\thetastar\|_\infty (1-\discount) \right \},
\end{align*}
where the final inequality follows since $\N \geq \unicon \frac{4^m
  \log(8 \NumEpoch \Dim/\delta)}{(1-\discount)^2}$. 

\subsection{Proof of Lemma~\ref{LemHighProbNoise}}
\label{SecProofLemHighProbNoise}

Our proof consists of two steps.  First, we prove by induction that
the moment-generating function of $\PnoiseA(\state, \action)$ is upper
bounded as
\begin{align}
 \label{EqnInductiveClaim}
  \log \Exs \left[ e^{s \PnoiseA(\state, \action)} \right] & \leq
  \frac{\Bou^2 s^2 \stepit{\titer-1}}{8} \qquad \mbox{for all $s \in
    \real$,}
\end{align}
uniformly over all state-action pairs $(\state, \action)$.  Combining
with the Chernoff bounding technique and the union bound, we find that
there is a universal constant $\unicon$ such that
\begin{align}
\mprob \left [ \|\pathitprime{\ell}\|_\infty \geq \unicon \Bou
  \sqrt{\stepit{\ell-1}} \sqrt{\log (8 \EpLength \NumEpoch
    \Dim/\delta)} \right] \leq \frac{\delta}{3 \EpLength \NumEpoch}.
\end{align}
Taking union bound over all $\EpLength$ iterations, we find that
\begin{align*}
  \frac{ 2 \sum_{\iter=1}^\EpLength \| \pathit{\ell}' \|_\infty}{1 +
    (1-\discount) \EpLength} + \|\pathit{\EpLength +1}'\|_\infty &
  \leq \frac{\unicon \Bou}{1 + (1-\discount) \EpLength} \sqrt{\log(8
    \EpLength \NumEpoch \Dim/\delta)} \left \{
  \sum_{\iter=1}^\EpLength \sqrt{\stepit{\iter-1}} +
  \sqrt{\stepit{\EpLength}} \right \},
\end{align*}
with probability at least $1 - \frac{\delta}{3 \NumEpoch}$.  From the
proof of Corollary 3 in the paper~\cite{Wai19CC}, we have
\begin{align*}
  \sum_{\iter=1}^\EpLength \sqrt{\stepit{\iter-1}} +
  \sqrt{\stepit{\EpLength}} & \leq \unicon \; \frac{ \sqrt{1 +
      (1-\discount) \titer}}{1-\discount},
\end{align*}
again for some universal constant $\unicon$.  Putting together the pieces
yields the claimed bound~\eqref{EqnHighProbNoise}.


\subsubsection{Proof of the claim~\eqref{EqnInductiveClaim}}

It remains to prove the bound~\eqref{EqnInductiveClaim}.  Recall that
the stochastic process $\{\pathitprime{\titer} \}_{\titer \geq 1}$
evolves according to the recursion $\pathitprime{\titer+1} =
(1-\stepit{\titer}) \pathitprime{\titer} + \stepit{\titer}
\VnoiseAIter{\titer}$, where
\begin{align*}
  \VnoiseAIter{\titer} & \defn \Hop(\mythetabar) - \HopK(\mythetabar)
  = \left \{ \Bellman(\mythetabar) - \Bellman(\thetastar) \right \} -
  \left \{ \EmpBellmanIter{\titer}(\mythetabar) -
  \EmpBellmanIter{\titer}(\thetastar) \right \}.
\end{align*}
Since the operator $\EmpBellmanIter{\titer}$ is $\discount$-contractive, we have
\begin{align*}
\| \EmpBellmanIter{\titer}(\mythetabar) -
\EmpBellmanIter{\titer}(\thetastar) \| & \leq \discount \|\mythetabar
- \thetastar\|_\infty \; \leq \; \Bou,
\end{align*}
where the final step uses the assumption that $\|\mythetabar -
\thetastar\|_\infty \leq \Bou$.  Moreover, we have
$\Exs[\HopK(\mythetabar)] = \Hop(\mythetabar)$, so that each
$\VnoiseAIter{\titer}$ is a zero-mean random matrix, with its entries
bounded in absolute value by $\Bou$.  Consequently, by standard
results on sub-Gaussian variables (cf. Chapter 2,~\cite{Wai19}), we
have
\begin{align}
  \label{EqnBaseSubGauss}
\log \Exs \left[ e^{s \VnoiseAIter{\titer}(\state, \action)} \right] &
\leq \frac{s^2 \Bou^2}{8} \qquad \mbox{for all $s \in \real$,}
\end{align}
valid for each state-action pair $(\state, \action)$.

We use this auxiliary result to prove the
claim~\eqref{EqnInductiveClaim} by induction.

\paragraph{Base case:}  The case $\titer = 1$ is trivial, since
$\pathitprime{1} = 0$.  Turning to the case $\titer = 2$, we have
$\pathitprime{2} = \stepit{1} \VnoiseAIter{1}$, and hence
\begin{align*}
  \log \Exs \left [e^{s \pathitprime{2}(\state, \action)} \right] & =
  \log \Exs \left [e^{s \stepit{1} \VnoiseAIter{1}(\state, \action)}
    \right] \; \leq \; \frac{s^2 \stepit{1}^2 \Bou^2}{8} \; \leq \;
  \frac{s^2 \stepit{1} \Bou^2}{8},
\end{align*}
where the final bound follows from the fact that $\stepit{1} =
\frac{1}{1 + (1-\discount)} \leq 1$.

\paragraph{Induction step:}
Next we assume that the claim~\eqref{EqnInductiveClaim} holds for some
iteration $\titer \geq 1$, and we verify that it holds at iteration
$\titer + 1$.  By definition of $\pathitprime{\titer+1}$ and the
independence of $\pathitprime{\titer}$ and $\VnoiseAIter{\titer}$, we
have
\begin{align*}
  \log \Exs \left[ e^{s \pathitprime{\titer+1} (\state, \action)}
    \right] & = \log \Exs \left[ e^{s (1-\stepit{\titer})
      \pathitprime{\titer} (\state, \action)} \right] + \log \Exs
  \left[ e^{s \stepit{\titer} \VnoiseAIter{\titer} (\state, \action)}
    \right] \\ & \leq \frac{s^2 (1-\stepit{\titer})^2
    \stepit{\titer-1} \Bou^2}{8} + \frac{s^2 \stepit{\titer}^2
    \Bou^2}{8}
\end{align*}
where the inequality makes use of the inductive assumption, and the
bound~\eqref{EqnBaseSubGauss}.  Recalling that $\stepit{\titer} =
\frac{1}{1 + (1-\discount) \titer}$, we have
\begin{align*}
(1-\stepit{\titer}) \stepit{\titer-1}  = \left( 1 - \frac{1}{1 +
    (1-\discount) \titer} \right) \frac{1}{1 + (1-\discount)
    (\titer-1)} & = \stepit{\titer}
  \underbrace{\left ( \frac{(1-\discount) +
  (1-\discount) (\titer-1)}{1 + (1-\discount)(\titer-1)}
    \right)}_{\leq 1}  \\
& \leq \stepit{\titer}
\end{align*}
Consequently, we have
\begin{align*}
\frac{s^2 (1-\stepit{\titer})^2 \stepit{\titer-1} \Bou^2}{8} +
\frac{s^2 \stepit{\titer}^2 \Bou^2}{8} & \leq \frac{s^2
  (1-\stepit{\titer}) \stepit{\titer} \Bou^2}{8} + \frac{s^2
  \stepit{\titer}^2 \Bou^2}{8} \; = \; \frac{s^2 \stepit{\titer}
  \Bou}{8},
\end{align*}
which verifies the claim~\eqref{EqnInductiveClaim} for $\titer + 1$.


\section{Auxiliary lemmas for Proposition~\ref{PropMinimax}}
\label{AppProofPropMinimax}

In this appendix, we collect the proofs of Lemmas~\ref{LemSharpEpoch}
and~\ref{LemSelfBounding}, used in the proof of
Proposition~\ref{PropMinimax}.


\subsection{Proof of Lemma~\ref{LemSharpEpoch}}
\label{AppProofLemSharpEpoch}

We begin by re-writing the recursion~\eqref{EqnHopJQvar} in a form
suitable for application of our results from past work~\cite{Wai19CC}.
Subtracting off the fixed point $\thetahat$ of $\HopJ$, we find that
\begin{align*}
\thetait{\titer+1} - \thetahat & = (1-\stepit{\titer})
(\thetait{\titer} - \thetahat) + \stepit{\titer} \left(
\HopJEmpIter{\titer}(\thetait{\titer}) -
\HopJEmpIter{\titer}(\thetahat) \right) + \stepit{\titer}
\left(\HopJEmpIter{\titer}(\thetahat) - \HopJ(\thetahat) \right).
\end{align*}
Note that the operator $\theta \mapsto \HopJEmpIter{\titer}(\theta) -
\HopJEmpIter{\titer}(\thetahat)$ is $\discount$-contractive and
monotonic so that Corollary 1 from Wainwright~\cite{Wai19CC} can be applied.
In applying this corollary, the effective noise term is given by
\begin{align*}
  \Eff_k & \defn \HopJEmpIter{\titer}(\thetahat) - \HopJ(\thetahat) \;
  = \; \left \{ \EmpBellmanIter{\titer}(\thetahat) -
  \EmpBellmanIter{\titer}(\thetabar) \right \} - \left \{
  \Bellman(\thetahat) - \Bellman(\thetabar) \right \}.
\end{align*}
Consequently, we have $\|\Eff_k\|_\infty \leq 2 \|\thetahat -
\thetabar\|_\infty$, a fact that is useful in bounding the effect of
these noise terms on the evolution.  By adapting Corollary 1 from the
paper~\cite{Wai19CC}, we have
\begin{align*}
\|\thetait{\EpLength + 1} - \thetahat\|_\infty & \leq \frac{2}{1 +
  (1-\discount) \EpLength} \left \{ \|\thetabar - \thetahat\|_\infty +
\sum_{\iter=1}^{\EpLength} \| \pathit{\iter} \|_\infty \right \} +
\|\pathit{\EpLength+1}\|_\infty,
\end{align*}
where the auxiliary stochastic process evolves as $\pathit{\titer} =
(1-\stepit{\titer-1}) \pathit{\titer-1} + \stepit{\titer-1}
\Eff_{\titer-1}$.  Following the same line of argument as in the proof
of Lemma~\ref{LemHighProbNoise} (see
Section~\ref{SecProofLemHighProbNoise}), we find that
\begin{align*}
  \|\thetait{\EpLength + 1} - \thetahat\|_\infty & \leq \unicon \left
  \{ \frac{ \|\thetabar - \thetahat\|_\infty}{1 + (1-\discount)
    \EpLength} + \frac{\|\thetabar -
    \thetahat\|_\infty}{(1-\discount)^{3/2} \sqrt{\EpLength}} \right
  \} \sqrt{\log(8 \Dim \NumEpoch/\delta)}
\end{align*}
with probability at least $1 - \frac{\delta}{2 \NumEpoch}$.  Here we
have used the fact that $\frac{\log(8 \Dim
  \NumEpoch/\delta)}{\EpLength} \leq 1$ by assumption.

Consequently, with the choice $\EpLength = \unicon_1 \frac{\log
  \left(\frac{8 \NumEpoch \Dim}{(1-\discount) \delta}
  \right)}{(1-\discount)^3}$, we are guaranteed that
\begin{align*}
  \|\thetait{\EpLength + 1} - \thetahat\|_\infty & \leq \frac{1}{4}
  \|\thetabar - \thetahat\|_\infty \leq \frac{1}{4} \|\thetabar -
  \thetastar\|_\infty + \frac{1}{4} \|\thetahat - \thetastar\|_\infty.
\end{align*}
with probability at least $1 - \frac{\delta}{2 \NumEpoch}$.


\subsection{Proof of Lemma~\ref{LemSelfBounding}}
\label{AppProofLemSelfBounding}

Note that $\thetahat$ is the fixed point of the operator
$\HopJ(\theta) \defn \Bellman(\theta) - \Bellman(\thetabar) +
\BellmanMixN(\thetabar)$, and hence can be seen as a fixed point of
the population Bellman operator defined with perturbed reward function
$\rtil$ with entries $\rtil(\state, \action) = \reward(\state,
\action) + \big[ \BellmanMixN(\thetabar) - \Bellman(\thetabar)
  \big](\state, \action)$.  The following lemma guarantees that this
perturbation is relatively small, where the reader should recall the
standard deviation $\sigma(\thetastar)$ that was previously
defined~\eqref{EqnBellmanVar}.
\begin{lemma}[Bounds on perturbed rewards]
\label{LemPertReward}
  For any matrix $\thetabar$ such that $\|\thetabar -
  \thetastar\|_\infty \leq \bou_\epochit$, we have
  \begin{align}
    \label{EqnPertReward}
\left|\rtil - \reward \right| & \preceq c \left ( \bou_\epochit
\onevec + \sigma(\thetastar) \right) \sqrt{\frac{\log(8 \NumEpoch
    \Dim/\delta)}{\N}} + \unicon' \|\thetastar\|_\infty \frac{\log (8
  \NumEpoch \Dim/\delta)}{\N} \onevec
\end{align}
with probability at least $1 - \frac{\delta}{8 \NumEpoch}$.
\end{lemma}

\noindent We also require a lemma that provides elementwise upper
bounds on the absolute difference $|\thetastar - \thetahat|$ in terms
of the absolute difference $|\rtil - \reward|$.  In order to state
these bounds, we follow the notation of Azar et
al.~\cite{AzaMunKap13}, letting $\PPol$ denote the linear operator
defined by the policy $\pol^*$ that is optimal with respect to
$\thetastar$, and similarly letting $\PhatPol$ denote the linear
operator defined by the policy $\polhat$ that is optimal with respect
to $\thetahat$.

\begin{lemma}[Elementwise bounds]
\label{LemElementwise}  
We have the elementwise upper bound:
\begin{align}
  \label{EqnElementwise}
\big| \thetastar - \thetahat \big| & \preceq \max \left \{ (\IdMat -
\discount \PPol)^{-1} \left| \rtil - \reward \right|, (\IdMat -
\discount \PhatPol)^{-1} \left|\rtil - \reward \right| \right \}.
\end{align}
\end{lemma}

\noindent Equipped with these lemmas, we now proceed to prove the
claim.  From the inequality~\eqref{EqnElementwise}, it suffices to
bound the elements of the two vectors $(\IdMat - \discount \PPol)^{-1}
\left| \rtil - \reward \right|$ and $(\IdMat - \discount
\PhatPol)^{-1} \left|\rtil - \reward \right|$.


\paragraph{Upper bounding $(\IdMat - \discount \PPol)^{-1}
  \left| \rtil - \reward \right|$:}

On one hand, from Lemma~\ref{LemPertReward} and the fact that the
matrix $(\IdMat - \discount \PPol)^{-1}$ has non-negative entries, we
have
\begin{align*}
(\IdMat - \discount \PPol)^{-1} \left| \rtil - \reward \right| &
  \preceq c \left ( \frac{\bou_\epochit}{1-\discount} + \|(\IdMat -
  \discount \PPol)^{-1} \sigma(\thetastar)\|_\infty \right)
  \sqrt{\frac{\log(8 \NumEpoch \Dim/\delta)}{\N}} \onevec + \unicon'
  \frac{\|\thetastar\|_\infty}{1-\discount} \frac{\log (8 \NumEpoch
    \Dim/\delta)}{\N} \onevec,
\end{align*}
where we have also used the fact that $\|(\IdMat - \discount
\PPol)^{-1} u\|_\infty \leq \frac{\|u\|_\infty}{1 - \discount}$ for
any vector $u$. Now we have
\begin{align*}
\|(\IdMat - \discount \PPol)^{-1} \sigma(\thetastar) \|_\infty &
\stackrel{(i)}{\leq} \frac{4 }{(1-\discount)^{3/2}}
\stackrel{(ii)}{\leq} \frac{4 \, (2^\epochit) }{1- \discount}
\bou_\epochit,
\end{align*}
where step (i) follows from Lemma 8 of Azar et al.~\cite{AzaMunKap13},
and step (ii) follows since $\bou_\epochit = \frac{1}{2^m}
\frac{1}{\sqrt{1-\discount}}$.  Similarly, we have
$\frac{\|\thetastar\|_\infty}{1-\discount} \leq
\frac{1}{(1-\discount)^2} \; \leq \; \frac{2^\epochit
  \bou_\epochit}{(1-\discount)^{3/2}}$.  Putting together the pieces
yields the elementwise bound
\begin{subequations}
  \begin{align}
\label{EqnFirstBound}    
(\IdMat - \discount \PPol)^{-1} \left| \rtil - \reward \right| &
\preceq \bou_\epochit \: \Phi(N, \epochit, \discount) \onevec
  \end{align}
where we define the non-negative scalar
\begin{align}
\label{EqnItchy}
\Phi(N, \epochit, \discount) & \defn \uniconprime \left
\{\frac{2^\epochit}{(1-\discount)} \sqrt{\LOGDNSQ} +
\frac{2^\epochit}{(1-\discount)^{3/2}} \LOGDNSQ \right \}
\end{align}
\end{subequations}
for a sufficiently large but universal constant $\uniconprime$.


\paragraph{Upper bounding $(\IdMat - \discount \PhatPol)^{-1}
  \left| \rtil - \reward \right|$:}

The only term that needs to be handled differently is the one
involving $\sigma(\thetastar)$.  Let $\sigma(\thetahat)$ denote the
variance under the transition function $\Pmat$ of the $Q$-function
$\thetahat$.  Again, by the results of Azar et al.~\cite{AzaMunKap13},
we are guaranteed that $\|(\IdMat - \discount \PhatPol)^{-1}
\sigma(\thetahat)\|_\infty \leq \frac{4}{(1-\discount)^{3/2}}$.
Moreover, we have $\sigma(\thetastar) \preceq \sigma(\thetahat) +
|\thetahat- \thetastar|$.  Combining the pieces, we are guaranteed to
have the elementwise bound
\begin{align}
  \label{EqnSecondBound}
(\IdMat - \discount \PhatPol)^{-1} \left| \rtil - \reward \right| &
  \preceq \bou_\epochit \; \Phi(\N, \epochit, \discount) \onevec +
  \unicon \frac{|\thetahat - \thetastar |}{1-\discount} \LOGDN
\end{align}
where the vector $\Phi$ was previously defined in
equation~\eqref{EqnItchy}.


\paragraph{Putting together the pieces:}

By combining the bounds~\eqref{EqnFirstBound}
and~\eqref{EqnSecondBound} with Lemma~\ref{LemElementwise}, we find
that
\begin{align*}
|\thetahat - \thetastar| & \preceq \bou_\epochit \; \Phi(\N, \epochit,
\discount) \onevec + \unicon \frac{|\thetahat - \thetastar
  |}{1-\discount} \LOGDN.
\end{align*}
Our choice of $\N$ ensures that $\frac{\unicon}{1-\discount} \LOGDN
\leq \frac{1}{2}$, so that we have established the upper bound
\mbox{$\|\thetahat - \thetastar\|_\infty \leq 2 \bou_\epochit \Phi(\N,
  \epochit, \discount)$.} Finally, returning to the
definition~\eqref{EqnItchy} of $\Phi$, we see that our choice of $\N$
ensures that \mbox{$\|\Phi(\N, \epochit, \discount)\|_\infty \leq
  \frac{1}{10}$,} so that the claim follows.
  

\subsubsection{Proof of Lemma~\ref{LemPertReward}}

Starting with the definition of $\rtil$ and adding and subtracting
terms, we obtain the bound
\begin{align*}
|\rtil - \reward| & = \left| \BellmanMixN(\thetabar) -
\Bellman(\thetabar) \right | \; \leq \; \left
|\left(\BellmanMixN(\thetabar) - \BellmanMixN(\thetastar) \right) -
\left(\Bellman(\thetabar) - \Bellman(\thetastar) \right) \right| +
\left| \BellmanMixN(\thetastar) - \Bellman(\thetastar) \right|.
\end{align*}
By definition, the random matrix $\BellmanMixN(\thetabar) -
\BellmanMixN(\thetastar)$ is the sum of $\N$ i.i.d. terms.  The
entries in each term are uniformly bounded by $\discount \|\thetabar -
\thetastar\|_\infty \leq \bou_\epochit$.  Consequently, by a
combination of Hoeffding's inequality and the union bound, we find
that
\begin{align*}
\left \| \left(\BellmanMixN(\thetabar) - \BellmanMixN(\thetastar)
\right) - \left(\Bellman(\thetabar) - \Bellman(\thetastar) \right)
\right \|_\infty & \leq 4 \bou_\epochit \sqrt{\LOGDNSQ}
\end{align*}
with probability at least $1 - \frac{\delta}{4 \NumEpoch}$.  Turning
to the term $|\BellmanMixN(\thetastar) - \Bellman(\thetastar)|$, by a
Bernstein inequality, we have
\begin{align*}
|\BellmanMixN(\thetastar) - \Bellman(\thetastar)| & \leq \unicon \left
\{ \sigma(\thetastar) \sqrt{\LOGDNSQ} + \|\thetastar\|_\infty \LOGDNSQ
\right \}.
\end{align*}
Combining the pieces yields the claim.


\subsubsection{Proof of Lemma~\ref{LemElementwise}}

In this proof, we make use of the function $\maxplus{u} = \max \{u,
0\}$, applied elementwise to a vector $u$.  Note that we have $|u| =
\max \{ \maxplus{u}, \; \maxplus{-u} \}$ by definition.  Using this
fact, it suffices to prove the two elementwise bounds:
\begin{align}
  \label{EqnFather}
\maxplus{\thetastar - \thetahat} \stackrel{(i)}{\preceq} (\IdMat -
\discount \PPol)^{-1} |\rtil - \reward|, \quad \mbox{and} \quad
\maxplus{\thetahat - \thetastar} \stackrel{(ii)}{\preceq} (\IdMat -
\discount \PhatPol)^{-1} \left|\rtil - \reward \right|.
\end{align}
Recall that $\thetastar$ and $\thetahat$ are the optimal $Q$-functions
for the reward functions $\reward$ and $\rtil$, respectively, with
corresponding optimal policies $\polopt$ and $\polhat$,
respectively. By this optimality, we have
\begin{align}
  \thetahat  = \rtil + \discount \PhatPol \thetahat \; \succeq \;
  \rtil + \discount \PPol \thetahat \quad \mbox{and} \quad
  \thetastar  = \reward + \discount \PPol \thetastar \succeq
  \reward + \discount \PhatPol \thetastar.
\end{align}

\paragraph{Proof of inequality~\eqref{EqnFather}(i):}
Using these relations, we can write
\begin{align*}
\thetastar - \thetahat = \left(\reward - \rtil \right) + \discount
\PPol \thetastar - \discount \PhatPol \thetahat & \preceq \left |\rtil -
\reward \right| + \discount \PPol (\thetastar - \thetahat) \\
& \preceq \left |\rtil - \reward \right| + \discount \PPol
\maxplus{\thetastar - \thetahat}
\end{align*}
where we have used the non-negativity of the entries of $\discount
\PPol$, and the fact that $\thetastar - \thetahat \preceq
\maxplus{\thetastar - \thetahat}$.  Since the RHS is non-negative,
this inequality implies that
\begin{align*}
\maxplus{\thetastar - \thetahat} & \preceq \left |\rtil - \reward
\right| + \discount \PPol \maxplus{\thetastar - \thetahat}
\end{align*}
Re-arranging and using the non-negativity of the entries of the matrix
$(\IdMat - \discount \PPol)^{-1}$, we find that $\maxplus{\thetastar -
  \thetahat} \preceq (\IdMat - \discount \PPol)^{-1} |\rtil -
\reward|$, as claimed in inequality~\eqref{EqnFather}(i).

\paragraph{Proof of inequality~\eqref{EqnFather}(ii):}
In the other direction, similar reasoning yields
\begin{align*}
\thetahat - \thetastar & = \left(\reward- \rtil \right) + \discount
\PhatPol \thetahat - \discount \PPol \thetastar \; \preceq \; \left
|\rtil - \reward \right| + \discount \PhatPol (\thetahat - \thetastar
) \preceq \; \left |\rtil - \reward \right| + \discount \PhatPol
\maxplus{\thetahat - \thetastar}
\end{align*}
and hence $\maxplus{\thetahat - \thetastar} \preceq (\IdMat -
\discount \PhatPol)^{-1} \left|\rtil - \reward \right|$, as claimed in
equation~\eqref{EqnFather}(ii).


\bibliographystyle{plain}


\begin{thebibliography}{10}

\bibitem{abbasi2018}
Y.~Abbasi-Yadkori, N.~Lazic, and C.~Szepesv{\'a}ri.
\newblock Regret bounds for model-free linear quadratic control.
\newblock {\em arXiv preprint arXiv:1804.06021}, 2018.

\bibitem{abbasi2011}
Y.~Abbasi-Yadkori and C.~Szepesv{\'a}ri.
\newblock Regret bounds for the adaptive control of linear quadratic systems.
\newblock In {\em Conference on Learning Theory}, pages 1--26, 2011.

\bibitem{abeille2017}
M.~Abeille and A.~Lazaric.
\newblock Improved regret bounds for {T}hompson sampling in linear quadratic
  control problems.
\newblock In {\em International {C}onfernce on {M}achine {L}earning}, pages
  1--9, 2018.

\bibitem{AgaJia17}
S.~Agrawal and R.~Jia.
\newblock Optimistic posterior sampling for reinforcement learning: worst-case
  regret bounds.
\newblock In {\em Advances in Neural Information Processing Systems}, pages
  1184--1194, 2017.

\bibitem{Aza11}
M.~G. Azar, R.~Munos, M.~Ghavamzadeh, and H.~J. Kappen.
\newblock Speedy {$Q$}-learning.
\newblock In {\em Neural Information Processing Systems}, pages 2411--2419,
  2011.

\bibitem{AzaMunKap13}
M.~G. Azar, R.~Munos, and H.~J. Kappen.
\newblock Minimax {P}{A}{C} bounds on the sample complexity of reinforcement
  learning with a generative model.
\newblock {\em Machine Learning}, 91:325--349, 2013.

\bibitem{AzaOsbMun17}
M.~G. Azar, I.~Osband, and R.~Munos.
\newblock Minimax regret bounds for reinforcement learning.
\newblock In {\em International {C}onference on {M}achine {L}earning}, 2017.

\bibitem{Bertsekas_dyn1}
D.~P. Bertsekas.
\newblock {\em Dynamic programming and stochastic control}, volume~1.
\newblock Athena Scientific, Belmont, MA, 1995.

\bibitem{BerTsi96}
D.~P. Bertsekas and J.~N. Tsitsiklis.
\newblock {\em Neuro-Dynamic Programming}.
\newblock Athena Scientific, 1st edition, 1996.

\bibitem{CohKorMan19}
A.~Cohen, T.~Koren, and Y.~Mansour.
\newblock Learning linear-quadratic regulators efficiently with only
  $\sqrt{T}$-regret.
\newblock Technical Report arXiv:1902.06223, arXiv, February 2019.

\bibitem{DefBacLac14}
A.~Defazio, F.~Bach, and S.~{L}acoste {J}ulien.
\newblock {S}{A}{G}{A}: {A} fast incremental gradient method with support for
  non-strongly convex composite objectives.
\newblock In {\em {N}{I}{P}{S} {C}onference}, 2014.

\bibitem{Du17}
S.~S. Du, J.~Chen, L.~Li, L.~Xiao, and D.~Zhou.
\newblock {S}tochastic variance reduction methods for policy evaluation.
\newblock Technical Report arxiv:1702.07944, {M}icrosoft {R}esearch, February
  2017.

\bibitem{EveMan03}
E.~Even-{D}ar and Y.~Mansour.
\newblock Learning rates for {$Q$}-learning.
\newblock {\em Journal of Machine Learning Research}, 5:1--25, 2003.

\bibitem{FazGeKakMes18}
M.~Fazel, R.~Ge, S.~Kakade, and M.~Mesbahi.
\newblock Global convergence of policy gradient methods for the linear
  quadratic regulator.
\newblock In {\em {I}nternational {C}onference on {M}achine {L}earning}, pages
  1466--1475, 2018.

\bibitem{JaaJorSin94}
T.~Jaakkola, M.~I. Jordan, and S.~P. Singh.
\newblock On the convergence of stochastic iterative dynamic programming
  algorithms.
\newblock {\em Neural Computation}, 6(6), November 1994.

\bibitem{JinZhuBubJor18}
C.~Jin, Z.~Allen-Zhu, S.~Bubeck, and M.~I. Jordan.
\newblock Is {$Q$}-learning provably efficient?
\newblock Technical report, arxiv, July 2018.

\bibitem{JohZha13}
R.~Johnson and T.~Zhang.
\newblock Accelerating stochastic gradient descent using predictive variance
  reduction.
\newblock In {\em {N}{I}{P}{S} {C}onference}, 2013.

\bibitem{KeaSin99}
M.~Kearns and S.~Singh.
\newblock Finite-sample convergence rates for {$Q$}-learning and indirect
  algorithms.
\newblock In {\em NIPS {C}onference}, 1999.

\bibitem{LatHut14}
T.~Lattimore and M.~Hutter.
\newblock Near-optimal {P}{A}{C} bounds for discounted {M}{D}{P}s.
\newblock {\em Theoretical {C}omputer {S}cience}, 558:125--143, 2014.

\bibitem{levine15}
S.~Levine, C.~Finn, T.~Darrell, and P.~Abbeel.
\newblock End-to-end training of deep visuomotor policies.
\newblock {\em Journal of Machine Learning Research}, 17(1):1334--1373, 2016.

\bibitem{MalPanBhaKhaBarWai19}
D.~Malik, A.~Panajady, K.~Bhatia, K.~Khamaru, P.~L. Bartlett, and M.~J.
  Wainwright.
\newblock Derivative-free methods for policy optimization: {G}uarantees for
  linear-quadratic systems.
\newblock In {\em AISTATS: Conference on AI and Statistics}, 2019.

\bibitem{ManTuRec19}
H.~Mania, S.~Tu, and B.~Recht.
\newblock Certainty equivalent control of {L}{Q}{R} is efficient.
\newblock Technical Report arXiv:1902.07826, arXiv, February 2019.

\bibitem{mnih15}
V.~Mnih et~al.
\newblock Human-level control through deep reinforcement learning.
\newblock {\em Nature}, 518(7540):529--533, February 2015.

\bibitem{Puterman05}
M.~L. Puterman.
\newblock {\em Markov decision processes: Discrete stochastic dynamic
  programming}.
\newblock Wiley, 2005.

\bibitem{SchRouBac17}
M.~Schmidt, N.~{L}e {R}oux, and F.~Bach.
\newblock Minimizing finite sums with the stochastic average gradient.
\newblock {\em Mathematical {P}rogramming}, 162:83--112, March 2017.

\bibitem{ShaZha13}
S.~Shalev-{S}hwartz and T.~Zhang.
\newblock Stochastic dual coordinate ascent methods for regularized loss
  minimization.
\newblock {\em {J}ournal of {M}achine {L}earning {R}esearch}, 14:567--599,
  2013.

\bibitem{Sid18b}
A.~Sidford, M.~Wang, C.~Wu, L.~Yang, and Y.~Ye.
\newblock Near-optimal time and sample complexities for solving {M}arkov
  decision processes with a generative model.
\newblock In {\em NeurIPS: Advances in Neural Information Processing Systems},
  2018.

\bibitem{Sid18a}
A.~Sidford, M.~Wang, X.~Wu, and Y.~Ye.
\newblock Variance reduced value iteration and faster algorithms for solving
  {M}arkov decision processes.
\newblock In {\em Symposium on {D}iscrete {A}lgorithms ({S}{O}{D}{A})}, 2018.

\bibitem{silver16}
D.~Silver et~al.
\newblock Mastering the game of {Go} with deep neural networks and tree search.
\newblock {\em Nature}, 529(7587):484--489, January 2016.

\bibitem{SutBar18}
R.~S. Sutton and A.~G. Barto.
\newblock {\em {R}einforcement {L}earning: {A}n {I}ntroduction}.
\newblock {M}{I}{T} {P}ress, Cambridge, {M}{A}, 2nd edition, 2018.

\bibitem{Sze97}
C.~Szepesv\'{a}ri.
\newblock The asymptotic convergence rate of {$Q$}-learning.
\newblock In {\em NIPS 10}, pages 1064--1070, 1997.

\bibitem{Sze09}
C.~Szepesv\'{a}ri.
\newblock {\em Algorithms for reinforcement learning}.
\newblock Morgan-{C}laypool, 2009.

\bibitem{tobin17}
J.~Tobin, R.~Fong, A.~Ray, J.~Schneider, W.~Zaremba, and P.~Abbeel.
\newblock Domain randomization for transferring deep neural networks from
  simulation to the real world.
\newblock In {\em Intelligent Robots and Systems (IROS)}, pages 23--30. IEEE,
  2017.

\bibitem{Tsi94}
J.~N. Tsitsiklis.
\newblock Asynchronous stochastic approximation and {$Q$}-learning.
\newblock {\em Machine Learning}, 16:185--202, 1994.

\bibitem{TuRec19}
S.~Tu and B.~Recht.
\newblock The gap between model-based and model-free methods on the linear
  quadratic regulator: {A}n asymptotic viewpoint.
\newblock Technical report, {U}{C} {B}erkeley, February 2019.

\bibitem{Wai19}
M.~J. Wainwright.
\newblock {\em High-dimensional statistics: A non-asymptotic viewpoint}.
\newblock Cambridge University Press, Cambridge, {U}{K}, 2019.

\bibitem{Wai19CC}
M.~J. Wainwright.
\newblock Stochastic approximation with cone-contractive operators: Sharp
  $\ell_\infty$-bounds for {Q}-learning.
\newblock Technical report, UC Berkeley, May 2019.
\newblock arxiv:1905.06265.

\bibitem{WatDay92}
C.~Watkins and P.~Dayan.
\newblock ${Q}$-learning.
\newblock {\em Machine Learning}, 8:279--292, 1992.

\end{thebibliography}


\end{document}